\documentclass[11pt]{article}
\usepackage{acl}

\usepackage{times}
\usepackage{latexsym}
\usepackage[T1]{fontenc}
\usepackage[utf8]{inputenc}
\usepackage{microtype}
\usepackage{inconsolata}
\usepackage{graphicx}
\usepackage{url}
\usepackage{booktabs}
\usepackage{multirow}
\usepackage{subcaption}
\usepackage{amsmath}
\usepackage{amssymb}
\usepackage{amsthm}
\usepackage{newtxmath}
\usepackage{mathtools}
\providecommand{\UseTaggingSocket}[1]{}
\providecommand{\SuspendTagging}[1]{}
\providecommand{\ResumeTagging}[1]{}
\usepackage{tcolorbox}
\usepackage{xspace}

\definecolor{darkblue}{rgb}{0, 0, 0.5}
\hypersetup{colorlinks=true, citecolor=darkblue, linkcolor=darkblue, urlcolor=darkblue}

\title{When Do Intrinsic Rewards Work for Code Reasoning? A Comprehensive Study}

\author{
  Xiaolong Jin\textsuperscript{1} \quad
  Xuandong Zhao\textsuperscript{2} \quad
  Wenbo Guo\textsuperscript{3} \quad
  Xiangyu Zhang\textsuperscript{1} \quad
  Dawn Song\textsuperscript{2} \\
  \textsuperscript{1}Purdue University \quad
  \textsuperscript{2}UC Berkeley \quad
  \textsuperscript{3}UC Santa Barbara
}

\begin{document}

\maketitle

\begin{abstract}
Reinforcement learning with verifiable rewards (RLVR) has driven substantial progress in large language model reasoning, but relies on ground-truth supervision that is costly or infeasible, especially in coding tasks.
Recent work addresses this by deriving rewards from a model's own signals, such as majority voting or confidence-based scores, achieving notable success on mathematical reasoning benchmarks.
However, code generation poses distinct challenges: programs are structurally complex, semantically equivalent solutions may differ syntactically, and verification typically requires execution.
Whether these intrinsic reward methods transfer effectively to code remains unexplored.
In this work, we present a systematic empirical study of intrinsic reward methods for code generation.
We conduct extensive experiments on LiveCodeBench, systematically evaluating representative certainty-based Reinforcement Learning from
Internal Feedback (RLIF) approaches under different training scenarios and hyperparameter settings.
Our experiments reveal that certainty-based methods yield early gains but inevitably collapse: models progressively shorten outputs and lose reasoning capability, with collapse speed sensitive to sample size and temperature. When used to initialize RLVR training, RLIF pre-training offers no significant improvement over training from scratch.
We also provide actionable recommendations for using intrinsic rewards for training code reasoning models.
Our study shows both the promise and limitations of intrinsic reward methods for code, informing future work on code models and agents.
\end{abstract}

\section{Introduction}
\label{sec:introduction}

Reinforcement learning with verifiable rewards (RLVR) is the standard approach for improving LLM reasoning, where models receive binary or scalar rewards based on whether their outputs match ground-truth answers.
This paradigm has driven substantial progress in mathematical reasoning, with models such as OpenAI's o1~\citep{openaiOpenAIO1System2024} and o3, DeepSeek-R1~\citep{guoDeepSeekR1IncentivizesReasoning2025}, Gemini~\citep{comaniciGemini25Pushing2025} and Qwen3~\citep{yangQwen3TechnicalReport2025} demonstrating strong performance through scaled supervised RLVR.
However, supervised RLVR faces a fundamental bottleneck: it requires access to ground-truth verification, which may be difficult to obtain, especially for coding tasks.

This supervision bottleneck has motivated growing interest in Reinforcement Learning from Internal Feedback (RLIF), which derives rewards from the model's own internal signals rather than external labels. Existing methods fall into two categories. Ensemble-based methods, such as TTRL~\citep{zuoTTRLTestTimeReinforcement2025}, use agreement across multiple rollouts as a proxy for correctness, assuming that self-consistent answers are more likely to be right. Certainty-based methods, such as EM-RL~\citep{agarwalUnreasonableEffectivenessEntropy2025} and RENT~\citep{prabhudesaiMaximizingConfidenceAlone2025}, reward outputs with high confidence or low entropy, reinforcing predictions the model is already certain about. These approaches have shown promising results, primarily on mathematical reasoning benchmarks.

The coding domain presents a compelling setting for applying RLIF. Code-focused LLMs have become essential tools in software development~\citep{huiQwen25CoderTechnicalReport2024,teamGLM45AgenticReasoning2025}, and code agents capable of multi-step tasks such as debugging, repository navigation, and feature implementation represent an increasingly important frontier~\citep{jimenezSWEbenchCanLanguage2024,yangSWEagentAgentComputerInterfaces2024,teamTraeAgentLLMbased2025}. However, training such agents poses significant challenges: trajectories are long, reward signals are sparse, comprehensive test suites are rarely available~\citep{teamGLM45AgenticReasoning2025}, and execution requires sandboxed environments. These challenges make code agents a natural application for RLIF, where intrinsic rewards could serve as lightweight filters to identify high-quality trajectories, provide dense intermediate feedback, or enable training when ground-truth test cases are incomplete.
Yet it remains unclear whether RLIF methods developed for mathematics transfer effectively to code. Mathematical reasoning typically produces short, unique answers that can be directly compared across samples. Code generation, by contrast, produces programs where semantically equivalent solutions may differ substantially in surface form, making it difficult for ensemble-based methods to define agreement. Verification also differs: mathematical correctness can often be checked via string matching, while code correctness requires execution against test cases~\citep{jain2025livecodebench}. Despite these fundamental differences, existing empirical studies of RLIF~\citep{he2026far} have focused almost exclusively on mathematics, leaving the code domain largely unexplored.
It is unknown which methods work, what failure modes arise, and under what conditions intrinsic rewards provide reliable signal for code generation.

In this paper, we present a systematic empirical study of RLIF methods for code generation.
We implement representative approaches from certainty-based categories and evaluate them on function-level code generation tasks using LiveCodeBench. Our study addresses three questions: (1) How do different RLIF methods perform on code generation compared to supervised baselines? (2) What failure modes emerge, and how do they differ from those observed in mathematical reasoning? (3) Under what conditions do intrinsic rewards provide reliable training signal?

Our comprehensive experiments on LiveCodeBench~\citep{jain2025livecodebench} reveal several critical insights into RLIF for code reasoning.
We observe minimal performance variance across different RLIF methods, with all eventually leading to entropy collapse.
Notably, when integrated with GRPO, RLIF provides no significant improvement over the RLVR.
While collapse appears inevitable, its speed is sensitive to hyperparameters, specifically sample size $n$ and temperature.
For test-time training, RLIF collapses slower than the main training.
Across all settings, we observe a consistent reduction in model output length, suggesting that RLIF primarily reinforces the model's confidence in shorter responses.
While this can yield early-stage performance gains, it fails to provide long-term improvements.
From these findings, we recommend prioritizing obtaining verifiable rewards, such as through environment synthesis, or restricting RLIF use to early training stages with moderate $n$ and high $temperature$ settings.

\section{Related Work}
\label{sec:related_work}
\paragraph{Reinforcement Learning with Verifiable Rewards}
RLVR trains language models using rewards derived from verifiable correctness, typically through comparison with ground-truth answers~\citep{lambertTulu3Pushing2025}. This approach has been central to recent advances in reasoning, with leading models achieving strong performance on mathematics and science benchmarks through scaled supervised RLVR~\citep{openaiOpenAIO1System2024,guoDeepSeekR1IncentivizesReasoning2025,yangQwen3TechnicalReport2025}.
However, this reliance on ground-truth verification limits applicability in domains where labels are expensive or unavailable.

\paragraph{Reinforcement Learning from Internal Feedback}
To address the supervision bottleneck, recent work has explored RLIF methods that derive rewards from the model's own outputs or internal signal rather than external labels. These methods fall into two main categories.
\textbf{Ensemble-based methods} construct rewards from agreement across multiple samples. TTRL~\citep{zuoTTRLTestTimeReinforcement2025} pioneered the use of majority voting across rollouts to create pseudo-labels, treating consensus as a proxy for correctness. This approach builds on the intuition from self-consistency decoding~\citep{wangSelfConsistencyImprovesChain2023} that agreement among diverse reasoning paths indicates higher likelihood of correctness.
Follow-up work has analyzed limitations of this approach, including reward hacking when the majority answer is incorrect~\citep{shafayatCanLargeReasoning2025}, and proposed extensions such as entropy-guided tree search for improved efficiency~\citep{liuETTRLBalancingExploration2025}
and semantic clustering for soft voting~\citep{zhangConsistentPathsLead2025}.
\textbf{Certainty-based methods} derive rewards from a single policy's confidence along a trajectory. Intuitor~\citep{intuitor} rewards outputs with high self-certainty measured via KL divergence from uniform distributions. EM-RL~\citep{agarwalUnreasonableEffectivenessEntropy2025} and RENT~\citep{prabhudesaiMaximizingConfidenceAlone2025} use negative entropy as a reward signal, encouraging low-entropy predictions. Other approaches reward based on raw probability~\citep{liConfidenceAllYou2025}, probability disparity between top tokens~\citep{vanniekerkPostTrainingLargeLanguage2025}, or attention patterns~\citep{kirulutaSelfSupervisedReinforcementLearning2025}. These methods share a common principle: they reinforce outputs the model is already confident about, effectively sharpening existing preferences.
While both categories have shown gains on mathematical reasoning tasks, recent analyses have identified failure modes including reward hacking, reduced exploration, and model collapse when confidence misaligns with correctness~\citep{shafayatCanLargeReasoning2025,agarwalUnreasonableEffectivenessEntropy2025}. A concurrent work provides theoretical analysis of intrinsic rewards and empirically studies their scaling limits on mathematical reasoning~\citep{zhangNoFreeLunch2025}. Our work complements this line of research by focusing specifically on the code domain.

\section{RLIF: A Unified Perspective}
\label{sec:theoretical_framework}
Reinforcement learning with verifiable rewards (RLVR) trains language models by providing binary feedback based on answer correctness. Given an input $x$ with ground-truth answer $a$, the model generates a response $y$, and the reward is defined as:
\begin{equation}
    R(x, y, a) =
    \begin{cases}
        1 & \text{if } y \text{ is equivalent to } a \\
        0 & \text{otherwise}
    \end{cases}
\end{equation}

This formulation requires external verification, either through exact matching, execution-based evaluation, or other correctness checks. In code generation, however, such verification often depends on test cases, which may be unavailable or unreliable in repository-level and agent-based settings.

RLIF methods address this limitation by replacing external verification with internally derived signals. These methods construct rewards from the model's own output distributions, operating under the hypothesis that model confidence correlates with response quality. In this work, we focus on four representative approaches: Self-Certainty, Token-Level Entropy, Trajectory-Level Entropy, and Probability Disparity. Despite varied formulations, all four share a common structure that we formalize below.

\begin{table}[t]
\centering
\caption{
Summary of certainty-based reward methods used to compute the intrinsic
reward $r(x, y)$ in Eq.~\ref{eq:general}. Each row specifies the local confidence
measure $g(p_t, y_t)$ and the global transformation $f(\cdot)$.
``Length Norm.'' denotes whether the reward is normalized by sequence
length $T$.
}
\label{tab:methods}
\resizebox{\columnwidth}{!}{%
\begin{tabular}{lccc}
\toprule
\textbf{Method} & \textbf{Local Measure} $g(p_t, y_t)$ & \textbf{Transformation} $f(z)$ & \textbf{Length Norm.} \\
\midrule
Self-Certainty     & $D_{\mathrm{KL}}(\mathcal{U} \| p_t)$ & $z$            & Yes \\
Token Entropy      & $-H(p_t)$                             & $z$            & Yes \\
Trajectory Entropy & $\log p_t(y_t)$                       & $z$            & Yes \\
Probability Disparity & $\max_{v} p_t(v) - \max_{v \neq v^*_t} p_t(v)$ & $z$ & Yes \\
\bottomrule
\end{tabular}
}
\vspace{-1em}
\end{table}

\subsection{General Formulation}

Consider a language model $\pi_\theta$ that generates a response $y = (y_1, \ldots, y_T)$ given input $x$. At each decoding step $t$, the model produces a probability distribution $p_t(\cdot) = \pi_\theta(\cdot \mid x, y_{<t})$ over the vocabulary $\mathcal{V}$. Certainty-based rewards assess how concentrated these distributions are, either across the full vocabulary or specifically at the generated tokens.

All four methods in Table~\ref{tab:methods} can be expressed under a general form:
\begin{equation}
    r(x, y) = f\left( \frac{1}{T} \sum_{t=1}^{T} g(p_t, y_t) \right)
    \label{eq:general}
\end{equation}
where $g(\cdot, \cdot)$ is a local confidence measure computed at each position and $f(\cdot)$ is a global transformation. The methods differ in their choices of $g$ and $f$, which determine what aspect of confidence is captured and how the final reward is scaled. Table~\ref{tab:methods} summarizes these choices.

\subsection{Method Descriptions}

\paragraph{Self-Certainty~\citep{intuitor}.} This method measures how much the predictive distribution diverges from uniformity. Specifically, it computes the KL divergence from the uniform distribution $\mathcal{U}$ over $\mathcal{V}$ to the predictive distribution $p_t$ at each step:
\begin{equation}
    r_{\mathrm{SC}}(x, y) = \frac{1}{T} \sum_{t=1}^{T} D_{\mathrm{KL}}(\mathcal{U} \| p_t)
\end{equation}
By definition, $D_{\mathrm{KL}}(\mathcal{U} \| p_t) = \log|\mathcal{V}| + \frac{1}{|\mathcal{V}|} \sum_{v \in \mathcal{V}} \log p_t(v)$, which increases as $p_t$ concentrates on fewer tokens.
A higher value indicates that the model concentrates its probability mass on a small subset of vocabulary items, yielding higher rewards for decisive predictions.

\paragraph{Token-Level Entropy~\citep{prabhudesaiMaximizingConfidenceAlone2025}.} Instead of comparing to a reference distribution, this method directly penalizes uncertainty by measuring the entropy of the predictive distribution:
\begin{equation}
    r_{\mathrm{TE}}(x, y) = -\frac{1}{T} \sum_{t=1}^{T} H(p_t)
\end{equation}
where $H(p_t) = -\sum_{v \in \mathcal{V}} p_t(v) \log p_t(v)$ is the Shannon entropy. Lower entropy corresponds to a more concentrated distribution. By negating the entropy, the method assigns higher rewards to confident predictions.

\paragraph{Trajectory-Level Entropy~\citep{agarwalUnreasonableEffectivenessEntropy2025}.} Unlike the previous two methods that evaluate the full distribution, this method considers only the probability assigned to the generated tokens:
\begin{equation}
    r_{\mathrm{TrajE}}(x, y) = \frac{1}{T} \sum_{t=1}^{T} \log p_t(y_t) = \frac{1}{T} \log \pi_\theta(y \mid x)
\end{equation}
This shows that the reward equals the length-normalized log-probability of the entire response. Responses that the model assigns a higher likelihood receive higher rewards.

\paragraph{Probability Disparity~\citep{liConfidenceAllYou2025}.} This method measures the gap between the most likely and second most likely tokens at each position:
\begin{equation}
    r_{\mathrm{PD}}(x, y) = \frac{1}{T} \sum_{t=1}^{T} \left[ \max_{v} p_t(v) - \max_{v \neq v^*_t} p_t(v) \right]
\end{equation}
where $v^*_t = \arg\max_v p_t(v)$ is the most likely token at step $t$. A larger disparity indicates that the model strongly prefers one token over all alternatives, reflecting high decision confidence.

\subsection{Comparison}

The four methods differ along two dimensions. The first is measurement scope: Self-Certainty and Token-Level Entropy evaluate the full predictive distribution at each step, capturing how spread out the probability mass is across all vocabulary items; Trajectory-Level Entropy focuses only on the selected token, measuring how much probability mass falls on the actual generation path; Probability Disparity considers only the top two candidates, quantifying how decisively the model prefers its first choice over the runner-up.

The second dimension is token dependence: Self-Certainty, Token-Level Entropy, and Probability Disparity compute rewards based solely on the distribution $p_t$, independent of which token was actually sampled; Trajectory-Level Entropy depends explicitly on the generated token $y_t$, rewarding sequences that align with the model's predictions.

These differences, while subtle in formulation, can lead to distinct training dynamics.
The following sections empirically investigate whether these variations translate into meaningful differences in code generation performance.

\begin{figure*}[t]
    \centering
    \includegraphics[width=\textwidth]{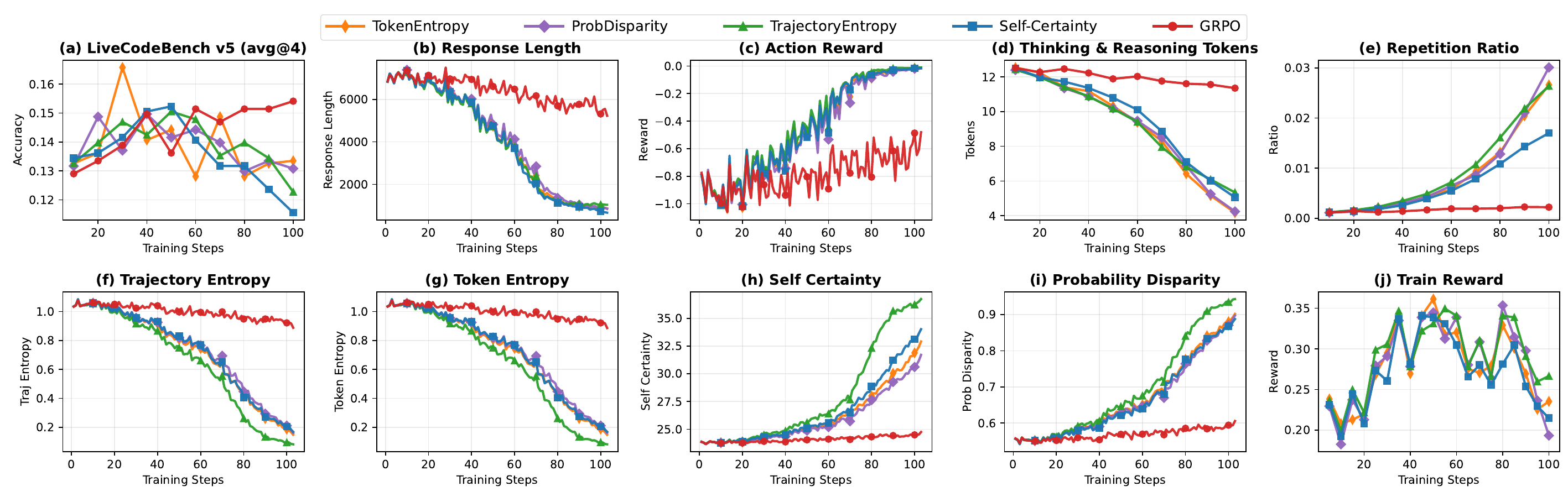}
    \caption{Training dynamics of all methods over 100 steps. (a) LiveCodeBench v5 avg@4 accuracy. (b) Average response length. (c) Action reward. (d) Thinking and reasoning token for validation response. (e) Repetition ratio for validation response. (j) Training Reward Score. (f) Trajectory entropy. (g) Token entropy. (h) Self-certainty. (i) Probability disparity. (j) Training reward score.}
    \label{fig:main_training_dynamics}
    \vspace{-0.3cm}
\end{figure*}

\begin{figure*}[t]
    \centering
    \includegraphics[width=\textwidth]{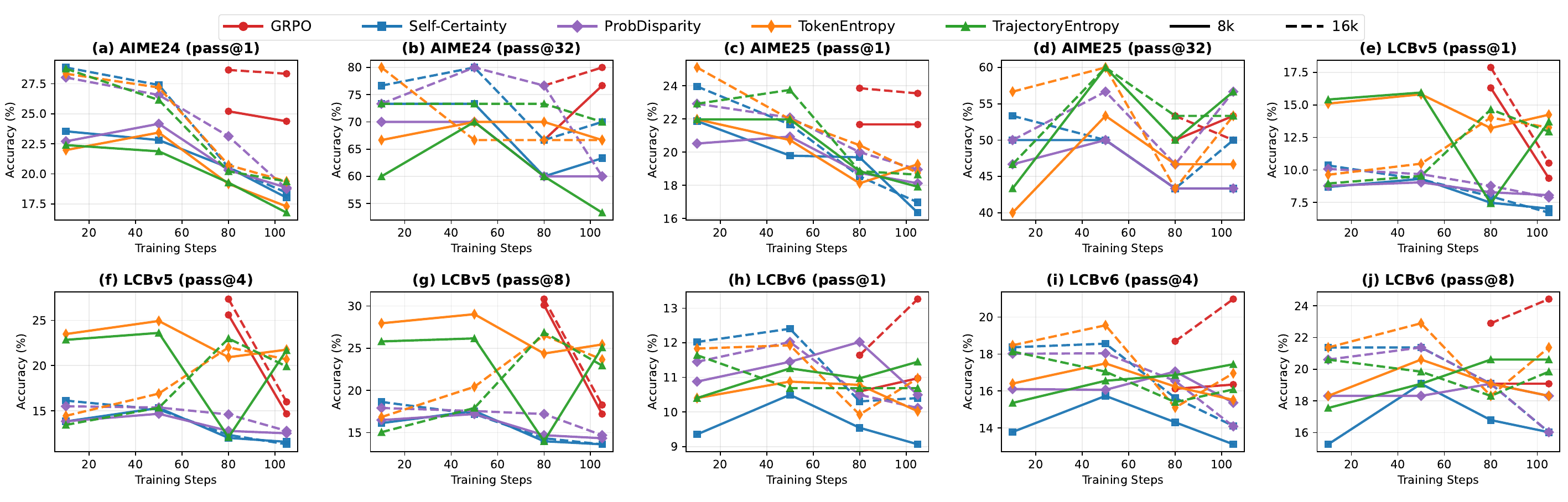}
\caption{Performance on mathematical reasoning (AIME24, AIME25) and code generation (LiveCodeBench v5, v6) benchmarks throughout training. Solid and dashed lines denote 8k and 16k maximum generation lengths in evaluation, respectively. Columns show pass@1, pass@K, and cons@32 metrics for each benchmark. cons@32 denotes majority-vote accuracy over 32 samples.}
    \label{fig:baseline_results}
    \vspace{-0.3cm}
\end{figure*}

\begin{figure*}[t]
    \centering
    \begin{subfigure}[b]{0.85\textwidth}
        \includegraphics[width=\textwidth]{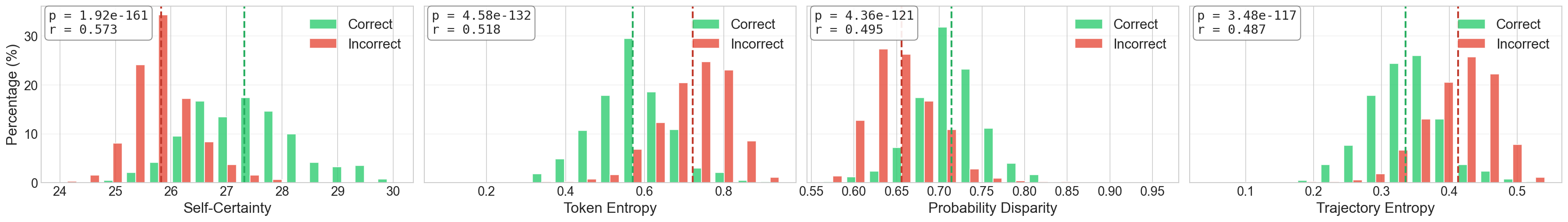}
        \caption{GRPO}
        \label{fig:certainty_grpo}
    \end{subfigure}
    \vspace{0.2cm}
    \begin{subfigure}[b]{0.85\textwidth}
        \includegraphics[width=\textwidth]{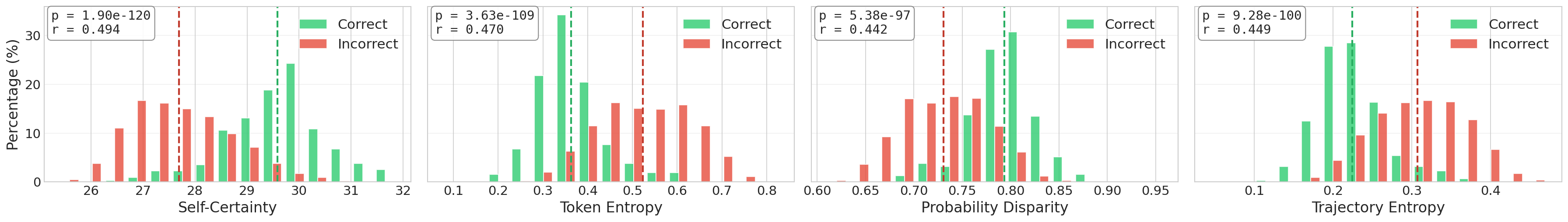}
        \caption{Self-Certainty}
        \label{fig:certainty_selfcert}
    \end{subfigure}
\caption{Distributions of four metrics (Self-Certainty, Token Entropy, Probability Disparity, Trajectory Entropy) for correct (green) and incorrect (red) solutions of checkpoints trained by GRPO and Self-Certainty on LiveCodeBench v5 at step 105. Dashed lines indicate class means; $r$ denotes rank-biserial correlation.}
    \vspace{-0.3cm}
    \label{fig:certainty_correctness}
\end{figure*}

\section{Experiments}
\label{sec:experiments}
\subsection{Setup}
\paragraph{Models.} We conduct experiments using DeepSeek-R1-Distill-Qwen-1.5B as the base model. This model is distilled from DeepSeek-R1~\citep{guoDeepSeekR1IncentivizesReasoning2025} using Qwen2.5-1.5B~\citep{yangQwen3TechnicalReport2025} as the backbone and fine-tuned on 800k high-quality reasoning samples.

\paragraph{Methods.} We compare GRPO~\citep{guoDeepSeekR1IncentivizesReasoning2025} against four certainty-based RLIF methods: Self-Certainty~\citep{intuitor}, Token-Level Entropy~\citep{prabhudesaiMaximizingConfidenceAlone2025}, Trajectory-Level Entropy~\citep{agarwalUnreasonableEffectivenessEntropy2025}, and Probability Disparity rewards~\citep{liConfidenceAllYou2025}. GRPO serves as our primary baseline as it represents the standard RLVR approach with verifiable rewards derived from external test cases. The four certainty-based RLIF methods instead compute intrinsic rewards from the model's own output distributions, requiring no external verification signals.

\paragraph{Training Data.} We adopt the training corpus from~\citep{archer}. The code training set comprises 6,753 programming problems drawn from competitive programming datasets~\citep{luo2025deepcoder,penedo2025codeforces,li2022competition}. These problems are augmented with extensive test cases to reduce false positives during reward computation.

\paragraph{Evaluation.} We evaluate all methods on both mathematical and coding benchmarks. For mathematics, we use six datasets: AIME24~\citep{maa2024aime}, AIME25~\citep{maa2025aime} and Minerva Math~\citep{minerva}. These benchmarks span competition-level problems and cover diverse mathematical topics, including algebra, geometry, number theory, and combinatorics.
For code generation, we use LiveCodeBench(LCB) v5 (2024.08.01--2025.02.01) and v6 (2025.02.01--2025.05.01)~\citep{jain2025livecodebench}, which focus on reasoning-intensive programming tasks.

Inference is performed using vLLM with a maximum output length of 8192 tokens and temperature set to 0.6 by default, $top_{p}=0.95$.
Following prior work, we report both avg@$K$ and pass@$K$ metrics to account for the high variance in reasoning model outputs.
For AIME24~\citep{maa2024aime}, AIME25~\citep{maa2025aime}, which contain relatively few problems, we set $K=32$ to obtain stable estimates. For LiveCodeBench v5, v6~\citep{jain2025livecodebench}, and Minerva Math~\citep{minerva}, we use $K=8$. Detailed metrics are shown in Appendix~\ref{app:evaluation metric}

\paragraph{Implementation Details.}
We implement all methods using the Verl framework. The training batch size is 64 with a mini-batch size of 32. We use a learning rate of $1 \times 10^{-6}$ without KL penalty, ppo epoch of 1. For each prompt, we sample 16 rollouts with temperature 1.0 and a maximum response length of 8192 tokens by default.

\subsection{Experiment Design}
\label{sec:experiment_design}
Our study addresses three research questions through corresponding experiments:

\paragraph{RQ1: Effectiveness.} How do RLIF methods perform compared to supervised RLVR?
We evaluate four certainty-based methods against GRPO on LiveCodeBench and AIME to measure both in-domain performance and cross-domain transfer, revealing whether gains reflect general reasoning improvements or domain-specific shortcuts.

\paragraph{RQ2: Failure Modes.} What causes RLIF to fail? We track training dynamics, including response length, entropy, reasoning token density, and repetition ratio to identify the mechanisms underlying performance degradation and model collapse.

\paragraph{RQ3: Reliability.} Under what conditions do intrinsic rewards provide reliable signal? We conduct four analyses: (i) certainty-correctness correlation to assess whether certainty can distinguish correct from incorrect solutions, (ii) bootstrapping GRPO from RLIF checkpoints to test whether early RLIF training provides useful initialization, (iii) test-time training on LiveCodeBench to evaluate RLIF in domain-specific settings, and (iv) ablations on rollout count and temperature to identify hyperparameter sensitivity.

\subsection{Main Results}
\label{sec:main_results}

As shown in Figure~\ref{fig:baseline_results}, we evaluate all methods on code generation benchmarks (LiveCodeBench v5, v6) and mathematical reasoning benchmarks (AIME24, AIME25) to assess both in-domain performance and cross-domain transfer under both 8k and 16k maximum output length settings.
On code generation benchmarks, GRPO with ground-truth rewards achieves consistent improvement, reaching 15.4\% avg@4 on LiveCodeBench at step 100. Certainty-based methods show competitive early performance, with TokenEntropy peaking at 16.6\% (step 30) and TrajectoryEntropy at 15.1\% (step 50), but all plateau or degrade thereafter. Self-Certainty declines most severely to 9.3\% by step 200. On LCBv5 (16k, step 80), GRPO leads with 17.9\%/27.3\%/30.8\% pass@1/4/8, while the best certainty methods TokenEntropy (14.0\%/22.0\%/26.5\%) and TrajectoryEntropy (14.7\%/23.0\%/26.9\%) trail consistently.

On mathematical reasoning benchmarks, GRPO demonstrates positive cross-domain transfer: AIME24 pass@1 remains stable around 28\% (16k) while cons@32 improves from 76.7\% to 80.0\% between steps 80 and 105. Certainty methods show the opposite trend, with ProbDisparity's AIME24 pass@32 dropping from 76.7\% to 60.0\% over the same interval, indicating that certainty-based optimization reinforces domain-specific patterns rather than general reasoning. Comparing 8k and 16k settings, GRPO consistently benefits from longer generation budgets (AIME24 pass@1: 25.2\% to 28.6\% at step 80), while this advantage vanishes for certainty methods at later checkpoints as models learn to generate truncated outputs.

\subsection{Training Dynamics and Failure Modes}
\label{sec:training_dynamics}

Figure~\ref{fig:main_training_dynamics} reveals the mechanisms underlying these performance differences. The most striking pattern is response length collapse: all certainty methods shrink from approximately 7,300 tokens (step 10) to 731--1,049 tokens (step 100), while GRPO maintains approximately 5,300 tokens. This reflects a degenerate incentive where shorter responses yield higher certainty scores regardless of problem complexity.

Length collapse is accompanied by entropy collapse and overconfidence. Trajectory entropy drops from 1.05 to 0.10--0.21 for certainty methods (GRPO: 0.92), self-certainty rises from 23.8 to 30.6--36.2 (GRPO: 24.5), and action reward approaches zero (GRPO: $-0.49$), indicating highly concentrated distributions without improved problem-solving. Reasoning token density degrades from 12.5 to 4.2--5.3 (GRPO: 11.4), showing that certainty optimization eliminates the extended reasoning chains needed for complex problems, while repetition ratio spikes from 0.001 to 0.017--0.030 (GRPO: 0.002).

These findings demonstrate that certainty-based methods provide a useful early training signal but cannot sustain continued training due to length collapse, entropy collapse, reasoning degradation, and severe repetition.

\begin{figure*}[htbp]
    \centering
    \includegraphics[width=\textwidth]{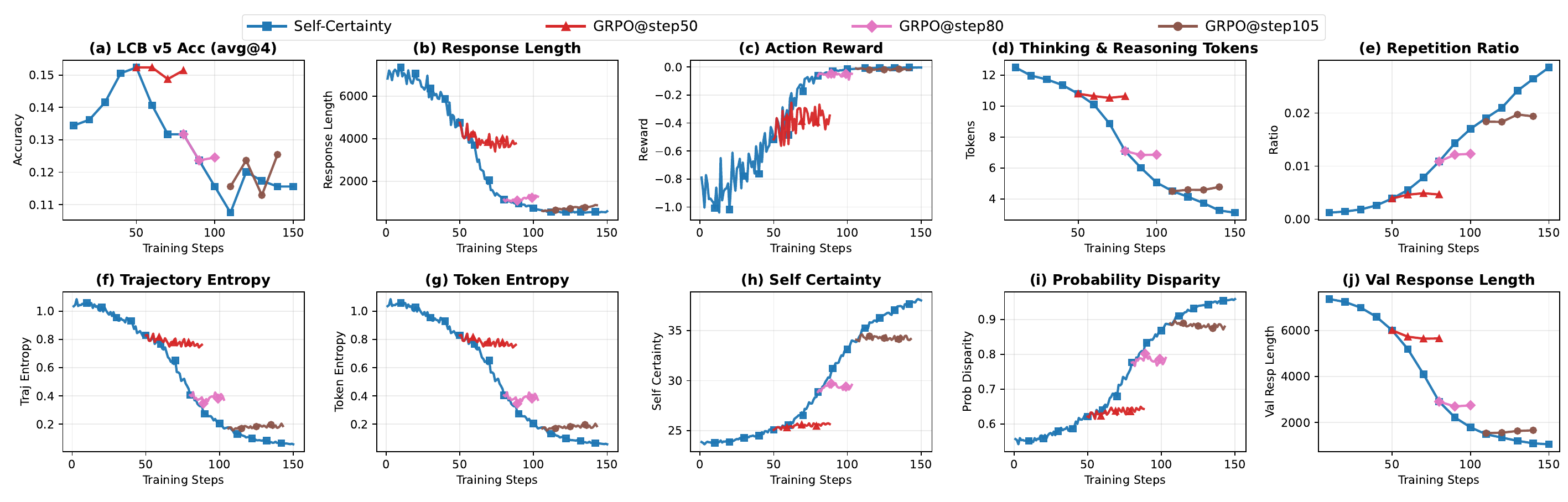}
\caption{Training dynamics when continuing GRPO training from Self-Certainty checkpoints at various training steps. Each GRPO run continues for 40 steps from the corresponding checkpoint. Metrics include (a) LiveCodeBench v5 accuracy, (b) response length, (c) action reward, (d) thinking \& reasoning tokens, (e) repetition ratio, (f) trajectory entropy, (g) token entropy, (h) self-certainty, (i) probability disparity, and (j) validation response length.}
    \label{fig:bootstrap_grpo}
    \vspace{-0.5cm}
\end{figure*}

\subsection{Certainty-Correctness Analysis}
\label{sec:certainty-correctness}
The previous experiments demonstrate that RLIF methods provide useful learning signals during early training but ultimately fail to match GRPO's performance due to entropy collapse and length compression. Yet a more fundamental question remains: can certainty reliably distinguish correct solutions from incorrect ones? We investigate this question by analyzing the distributions of certainty metrics for correct versus incorrect solutions on LiveCodeBench v5.
We evaluate GRPO and Self-Certainty checkpoints at step 105, generating 8 responses per problem. We quantify distributional separation using the rank-biserial correlation $r$ from the Mann-Whitney U test, where higher values indicate that correct solutions exhibit systematically higher certainty than incorrect ones.
As shown in Figure~\ref{fig:certainty_correctness}, the GRPO-trained model maintains stronger certainty-correctness correlations across all metrics compared to the Self-Certainty-trained model.
Examining the distributions more closely reveals the underlying mechanism. RLIF training shifts certainty scores rightward for both correct and incorrect solutions, meaning the model becomes uniformly more ``confident'' without becoming more accurate. This uniform inflation shrinks the gap between the two classes and undermines the discriminative power of certainty as a reward signal. Additional comparisons across all methods are provided in Appendix~\ref{app:certainty-correctness}.

\begin{figure*}[t]
    \centering
    \includegraphics[width=0.85\textwidth]{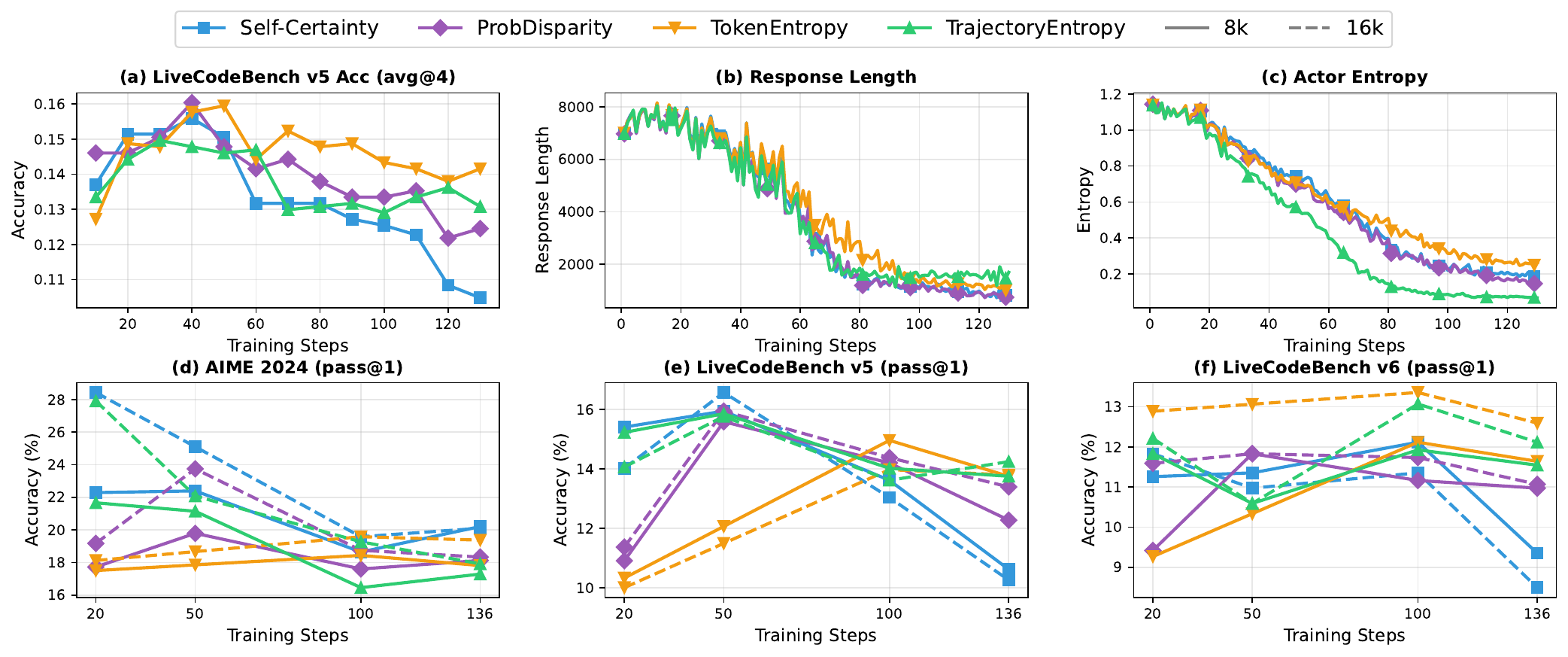}
    \caption{Test-time training on LiveCodeBench v5 by four RLIF methods. Top row: training dynamics including in-domain accuracy (avg@4), response length, and actor entropy. Bottom row: evaluation performance (pass@1) on AIME 2024, LCB v5, and LCB v6 across training steps. Solid lines denote 8k context; dashed lines denote 16k context.}
    \label{fig:lcb-main}
    \vspace{-0.5cm}
\end{figure*}

\subsection{Resuming RLVR from RLIF Checkpoints}
\label{sec:bootstrap}
The analysis above establishes that RLIF methods alone are insufficient for code generation. However, a practical question remains: can RLIF pre-training provide useful initialization for subsequent GRPO training?
This hybrid approach would reduce the test case construction cost during early training stage while still leveraging ground-truth rewards for final performance gains.
We investigate this question by taking checkpoints from certainty-trained models at various steps and continuing training with GRPO for 40 additional steps.
For Self-Certainty, we examine checkpoints at steps 10, 50, 80, and 105.
For ProbDisparity, TokenEntropy, and TrajectoryEntropy, we examine checkpoints at steps 50 and 105.

Figure~\ref{fig:bootstrap_grpo} presents the results.
Across all four methods, we observe a consistent pattern: early checkpoints bootstrap GRPO to performance levels matching GRPO trained from scratch, while later checkpoints yield substantially degraded performance that GRPO cannot recover.
For Self-Certainty, GRPO initialized from the step 50 checkpoint achieves 0.151 accuracy on LiveCodeBench v5, matching GRPO trained from scratch (0.154).
Later initializations show progressive degradation: step 80 reaches 0.125, and step 105 yields 0.120.
The pattern is consistent across other methods. GRPO@step50 outperforms GRPO@step105 for Probability Disparity, Token Entropy, and TrajectoryEntropy, shown in Figure~\ref{fig:bootstrap_grpo_probdisparity},\ref{fig:bootstrap_grpo_tokenentropy} and\ref{fig:bootstrap_grpo_trajectoryentropy}.
To understand why late checkpoints fail, we examine training dynamics in Figure~\ref{fig:bootstrap_grpo}.
Extended certainty training induces capability collapse along multiple dimensions: response length drops from 7,349 tokens to under 1,000 by step 90, reasoning-related tokens fall from 12 to below 5, and trajectory entropy collapses from 1.06 to 0.14 by step 105.
This collapse in entropy removes the variance required for the optimization of the policy gradient, making subsequent GRPO training ineffective, and the degradation proves difficult to reverse even after 40 additional steps of GRPO.
These results suggest a practical training strategy: certainty-based methods can serve as a test-case warm-up phase, provided the transition to GRPO occurs before capability collapse

\subsection{Test-Time Training on LiveCodeBench}
\label{exp:Test-Time Training on LCB}

A natural question is whether certainty-based methods perform better when trained directly on the target domain, where obtaining verifiable rewards may be impractical. We investigate this through \textit{test-time training}, training all self-RL methods on LiveCodeBench v5 (279 problems) for 3 epochs with default hyperparameters, and evaluating on LCB v5 (in-domain), LCB v6 (out-of-domain), and AIME.

Compared to training on diverse datasets in Section~\ref{sec:main_results}, self-RL methods degrade more slowly on this smaller domain-specific data. Response length collapses around step 70 rather than immediately, and Trajectory-Entropy achieves strong early performance, reaching 80\% pass@32 on AIME 2024 at step 50. However, this early promise diminishes with continued training. As shown in Figure~\ref{fig:lcb-main}(e), Self-Certainty's pass@1 on LCB v5 drops from 16\% to 10\%, while Token-Entropy remains relatively stable at 14\%. This degradation pattern is consistent across all RLIF variants and persists at higher pass@$k$. Probability Disparity shows additional instability, developing severe repetition (ratio 0.18 vs.\ under 0.05 for others).
Extending context length to 16k improves early-stage performance before models collapse. By step 136, this advantage largely disappears across methods, suggesting that longer context delays but does not prevent degradation. Overall, test-time training slows the decline of RLIF methods but does not resolve their fundamental instability.

\subsection{Ablation Studies}
\label{subsec:ablation}

We conduct ablation studies on rollout count $N \in \{8, 12, 16\}$, sampling temperature $\tau \in \{0.8, 1.0, 1.2\}$, KL coefficient $\beta_{\text{KL}} \in \{0, 0.005\}$, and PPO epochs $E \in \{1, 3\}$ for all methods. Each parameter is varied independently with others fixed at default values ($N=16$, $\tau=1.0$, $\beta_{\text{KL}}=0$, $E=1$). Detailed training dynamics for all configurations are provided in Appendix~\ref{app:ablation}.

\paragraph{Scaling across Model Sizes.}
The experiments above focus on DeepSeek-R1-Distill-Qwen-1.5B with an 8k training context. To assess whether these findings generalize, we extend the comparison to three additional configurations: R1-Distill-1.5B with 16k training context, Qwen3-4B (16k), and Qwen2.5-Coder-7B-Instruct (8k). We report training dynamics, evaluation curves, and KL penalty ablations across all configurations in Appendix~\ref{app:scaling}. These scaling experiments reveal that model size modulates the severity and onset of collapse, providing further insight into the conditions under which intrinsic rewards remain effective.

\section{Conclusion}
We presented a systematic empirical study of intrinsic reward methods for code generation. By evaluating four certainty-based RLIF approaches on LiveCodeBench, we identified a confidence-correctness trade-off: early training benefits from uncertainty reduction, while prolonged training risks reinforcing incorrect solutions. We characterized distinct failure modes across methods and identified conditions under which intrinsic rewards provide reliable signal.
Our findings suggest several directions for future work. 
First, developing semantic equivalence metrics for code could enable ensemble-based methods, which have shown strong results in mathematical reasoning. 
Second, hybrid approaches that combine intrinsic rewards with sparse ground-truth supervision may mitigate the failure modes we observe while retaining the sample efficiency benefits of intrinsic signals. 
Finally, adapting intrinsic signals to provide process rewards for code agents could guide intermediate reasoning steps in multi-stage workflows where outcome-level feedback is delayed or unavailable.

\clearpage

\section*{Ethics Statement}
This paper presents an empirical study of intrinsic reward methods for code generation. Our goal is to advance understanding of self-supervised training for code models.
There are many potential societal consequences of our work, such as potential misuse for generating malicious code.

\section*{Limitation}
Our study has several limitations.
First, we focus on function-level code generation using LiveCodeBench, which consists of self-contained algorithmic problems.
Real-world code generation often involves repository-level context, multi-file dependencies, and longer time horizons.
Whether our findings generalize to these more complex settings remains an open question.
Second, we evaluate only certainty-based RLIF intrinsic reward methods. Ensemble-based methods such as majority voting are difficult to apply to code because defining semantic equivalence without execution is non-trivial.
Developing effective equivalence metrics for code and evaluating ensemble-based methods is left for future work.
Third, we do not explore hybrid approaches that combine intrinsic rewards with sparse ground-truth supervision. Such combinations may mitigate the failure modes we identify.

\nocite{jimenezSWEbenchCanLanguage2024,yangSWEagentAgentComputerInterfaces2024,lambertTulu3Pushing2025,openaiOpenAIO1System2024,guoDeepSeekR1IncentivizesReasoning2025,yangQwen3TechnicalReport2025,leCodeRLMasteringCode2022,dou2024stepcoder,shojaeeExecutionbasedCodeGeneration2023,jiang2025coderl+,archer,chen2021evaluating}
\bibliography{colm_citation}

@inproceedings{
agarwalUnreasonableEffectivenessEntropy2025,
title={The Unreasonable Effectiveness of Entropy Minimization in {LLM} Reasoning},
author={Shivam Agarwal and Zimin Zhang and Lifan Yuan and Jiawei Han and Hao Peng},
booktitle={The Thirty-ninth Annual Conference on Neural Information Processing Systems},
year={2025},
url={https://openreview.net/forum?id=UfFTBEsLgI}
}

@article{comaniciGemini25Pushing2025,
  title={Gemini 2.5: Pushing the frontier with advanced reasoning, multimodality, long context, and next generation agentic capabilities},
  author={Comanici, Gheorghe and Bieber, Eric and Schaekermann, Mike and Pasupat, Ice and Sachdeva, Noveen and Dhillon, Inderjit and Blistein, Marcel and Ram, Ori and Zhang, Dan and Rosen, Evan and others},
  journal={arXiv preprint arXiv:2507.06261},
  year={2025}
}

@article{guoDeepSeekR1IncentivizesReasoning2025,
  title={Deepseek-r1: Incentivizing reasoning capability in llms via reinforcement learning},
  author={Guo, Daya and Yang, Dejian and Zhang, Haowei and Song, Junxiao and Zhang, Ruoyu and Xu, Runxin and Zhu, Qihao and Ma, Shirong and Wang, Peiyi and Bi, Xiao and others},
  journal={arXiv preprint arXiv:2501.12948},
  year={2025}
}

@inproceedings{
jimenezSWEbenchCanLanguage2024,
title={{SWE}-bench: Can Language Models Resolve Real-world Github Issues?},
author={Carlos E Jimenez and John Yang and Alexander Wettig and Shunyu Yao and Kexin Pei and Ofir Press and Karthik R Narasimhan},
booktitle={The Twelfth International Conference on Learning Representations},
year={2024},
url={https://openreview.net/forum?id=VTF8yNQM66}
}

@article{kirulutaSelfSupervisedReinforcementLearning2025,
  title={A Self-Supervised Reinforcement Learning Approach for Fine-Tuning Large Language Models Using Cross-Attention Signals},
  author={Kiruluta, Andrew and Lemos, Andreas and Burity, Priscilla},
  journal={arXiv preprint arXiv:2502.10482},
  year={2025}
}

@inproceedings{
lambertTulu3Pushing2025,
title={Tulu 3: Pushing Frontiers in Open Language Model Post-Training},
author={Nathan Lambert and Jacob Morrison and Valentina Pyatkin and Shengyi Huang and Hamish Ivison and Faeze Brahman and Lester James Validad Miranda and Alisa Liu and Nouha Dziri and Xinxi Lyu and Yuling Gu and Saumya Malik and Victoria Graf and Jena D. Hwang and Jiangjiang Yang and Ronan Le Bras and Oyvind Tafjord and Christopher Wilhelm and Luca Soldaini and Noah A. Smith and Yizhong Wang and Pradeep Dasigi and Hannaneh Hajishirzi},
booktitle={Second Conference on Language Modeling},
year={2025},
url={https://openreview.net/forum?id=i1uGbfHHpH}
}

@inproceedings{leCodeRLMasteringCode2022,
title={Code{RL}: Mastering Code Generation through Pretrained Models and Deep Reinforcement Learning},
author={Hung Le and Yue Wang and Akhilesh Deepak Gotmare and Silvio Savarese and Steven Hoi},
booktitle={Advances in Neural Information Processing Systems},
editor={Alice H. Oh and Alekh Agarwal and Danielle Belgrave and Kyunghyun Cho},
year={2022},
url={https://openreview.net/forum?id=WaGvb7OzySA}
}

@article{liConfidenceAllYou2025,
  title={Confidence Is All You Need: Few-Shot RL Fine-Tuning of Language Models},
  author={Li, Pengyi and Skripkin, Matvey and Zubrey, Alexander and Kuznetsov, Andrey and Oseledets, Ivan},
  journal={arXiv preprint arXiv:2506.06395},
  year={2025}
}

@article{liuETTRLBalancingExploration2025,
  title={Ettrl: Balancing exploration and exploitation in llm test-time reinforcement learning via entropy mechanism},
  author={Liu, Jia and He, ChangYi and Lin, YingQiao and Yang, MingMin and Shen, FeiYang and Liu, ShaoGuo},
  journal={arXiv preprint arXiv:2508.11356},
  year={2025}
}

@article{openaiOpenAIO1System2024,
  title = {{{OpenAI}} O1 {{System Card}}},
  author = {OpenAI},
  journal={arXiv preprint arXiv:2412.16720},
  year={2024}
}

@article{prabhudesaiMaximizingConfidenceAlone2025,
  title={Maximizing Confidence Alone Improves Reasoning},
  author={Prabhudesai, Mihir and Chen, Lili and Ippoliti, Alex and Fragkiadaki, Katerina and Liu, Hao and Pathak, Deepak},
  journal={arXiv preprint arXiv:2505.22660},
  year={2025}
}

@article{shafayatCanLargeReasoning2025,
  title={Can Large Reasoning Models Self-Train?},
  author={Shafayat, Sheikh and Tajwar, Fahim and Salakhutdinov, Ruslan and Schneider, Jeff and Zanette, Andrea},
  journal={arXiv preprint arXiv:2505.21444},
  year={2025}
}

@article{
shojaeeExecutionbasedCodeGeneration2023,
title={Execution-based Code Generation using Deep Reinforcement Learning},
author={Parshin Shojaee and Aneesh Jain and Sindhu Tipirneni and Chandan K. Reddy},
journal={Transactions on Machine Learning Research},
issn={2835-8856},
year={2023},
url={https://openreview.net/forum?id=0XBuaxqEcG},
note={}
}

@article{vanniekerkPostTrainingLargeLanguage2025,
  title={Post-training large language models via reinforcement learning from self-feedback},
  author={Niekerk, Carel and Vukovic, Renato and Ruppik, Benjamin Matthias and Lin, Hsien-chin and Ga{\v{s}}i{\'c}, Milica},
  journal={arXiv preprint arXiv:2507.21931},
  year={2025}
}

@inproceedings{wangSelfConsistencyImprovesChain2023,
title={Self-Consistency Improves Chain of Thought Reasoning in Language Models},
author={Xuezhi Wang and Jason Wei and Dale Schuurmans and Quoc V Le and Ed H. Chi and Sharan Narang and Aakanksha Chowdhery and Denny Zhou},
booktitle={The Eleventh International Conference on Learning Representations },
year={2023},
url={https://openreview.net/forum?id=1PL1NIMMrw}
}

@article{yangQwen3TechnicalReport2025,
  title={Qwen3 technical report},
  author={Yang, An and Li, Anfeng and Yang, Baosong and Zhang, Beichen and Hui, Binyuan and Zheng, Bo and Yu, Bowen and Gao, Chang and Huang, Chengen and Lv, Chenxu and others},
  journal={arXiv preprint arXiv:2505.09388},
  year={2025}
}

@inproceedings{
yangSWEagentAgentComputerInterfaces2024,
title={{SWE}-agent: Agent-Computer Interfaces Enable Automated Software Engineering},
author={John Yang and Carlos E Jimenez and Alexander Wettig and Kilian Lieret and Shunyu Yao and Karthik R Narasimhan and Ofir Press},
booktitle={The Thirty-eighth Annual Conference on Neural Information Processing Systems},
year={2024},
url={https://openreview.net/forum?id=mXpq6ut8J3}
}

@inproceedings{
zhangConsistentPathsLead2025,
title={Consistent Paths Lead to Truth: Self-Rewarding Reinforcement Learning for {LLM} Reasoning},
author={Kongcheng Zhang and QI YAO and Shunyu Liu and Yingjie Wang and Baisheng Lai and Jieping Ye and Mingli Song and Dacheng Tao},
booktitle={The Thirty-ninth Annual Conference on Neural Information Processing Systems},
year={2025},
url={https://openreview.net/forum?id=ckW70ls93V}
}

@article{zhangNoFreeLunch2025,
  title={No Free Lunch: Rethinking Internal Feedback for LLM Reasoning},
  author={Zhang, Yanzhi and Zhang, Zhaoxi and Guan, Haoxiang and Cheng, Yilin and Duan, Yitong and Wang, Chen and Wang, Yue and Zheng, Shuxin and He, Jiyan},
  journal={arXiv preprint arXiv:2506.17219},
  year={2025}
}

@inproceedings{
intuitor,
title={Learning to Reason without External Rewards},
author = {Zhao, Xuandong and Kang, Zhewei and Feng, Aosong and Levine, Sergey and Song, Dawn},
booktitle={The Fourteenth International Conference on Learning Representations},
year={2026},
url={https://openreview.net/forum?id=OU9nFEYR2M}
}

@inproceedings{
zuoTTRLTestTimeReinforcement2025,
title={{TTRL}: Test-Time Reinforcement Learning},
author={Yuxin Zuo and Kaiyan Zhang and Li Sheng and Shang Qu and Ganqu Cui and Xuekai Zhu and Haozhan Li and Yuchen Zhang and Xinwei Long and Ermo Hua and Biqing Qi and Youbang Sun and Zhiyuan Ma and Lifan Yuan and Ning Ding and Bowen Zhou},
booktitle={The Thirty-ninth Annual Conference on Neural Information Processing Systems},
year={2025},
url={https://openreview.net/forum?id=VuVhgEiu20}
}

@article{huiQwen25CoderTechnicalReport2024,
  title={Qwen2. 5-coder technical report},
  author={Hui, Binyuan and Yang, Jian and Cui, Zeyu and Yang, Jiaxi and Liu, Dayiheng and Zhang, Lei and Liu, Tianyu and Zhang, Jiajun and Yu, Bowen and Lu, Keming and others},
  journal={arXiv preprint arXiv:2409.12186},
  year={2024}
}

@inproceedings{
jain2025livecodebench,
title={LiveCodeBench: Holistic and Contamination Free Evaluation of Large Language Models for Code},
author={Naman Jain and King Han and Alex Gu and Wen-Ding Li and Fanjia Yan and Tianjun Zhang and Sida Wang and Armando Solar-Lezama and Koushik Sen and Ion Stoica},
booktitle={The Thirteenth International Conference on Learning Representations},
year={2025},
url={https://openreview.net/forum?id=chfJJYC3iL}
}

@article{teamGLM45AgenticReasoning2025,
  title={Glm-4.5: Agentic, reasoning, and coding (arc) foundation models},
  author={Zeng, Aohan and Lv, Xin and Zheng, Qinkai and Hou, Zhenyu and Chen, Bin and Xie, Chengxing and Wang, Cunxiang and Yin, Da and Zeng, Hao and Zhang, Jiajie and others},
  journal={arXiv preprint arXiv:2508.06471},
  year={2025}
}

@article{teamTraeAgentLLMbased2025,
  title={Trae agent: An llm-based agent for software engineering with test-time scaling},
  author={Gao, Pengfei and Tian, Zhao and Meng, Xiangxin and Wang, Xinchen and Hu, Ruida and Xiao, Yuanan and Liu, Yizhou and Zhang, Zhao and Chen, Junjie and Gao, Cuiyun and others},
  journal={arXiv preprint arXiv:2507.23370},
  year={2025}
}

@article{dou2024stepcoder,
  title={Stepcoder: Improve code generation with reinforcement learning from compiler feedback},
  author={Dou, Shihan and Liu, Yan and Jia, Haoxiang and Xiong, Limao and Zhou, Enyu and Shen, Wei and Shan, Junjie and Huang, Caishuang and Wang, Xiao and Fan, Xiaoran and others},
  journal={arXiv preprint arXiv:2402.01391},
  year={2024}
}

@article{jiang2025coderl+,
  title={CodeRL+: Improving Code Generation via Reinforcement with Execution Semantics Alignment},
  author={Jiang, Xue and Dong, Yihong and Liu, Mengyang and Deng, Hongyi and Wang, Tian and Tao, Yongding and Cao, Rongyu and Li, Binhua and Jin, Zhi and Jiao, Wenpin and others},
  journal={arXiv preprint arXiv:2510.18471},
  year={2025}
}

@article{archer,
  title={Stabilizing knowledge, promoting reasoning: Dual-token constraints for rlvr},
  author={Wang, Jiakang and Liu, Runze and Zhang, Fuzheng and Li, Xiu and Zhou, Guorui},
  journal={arXiv preprint arXiv:2507.15778},
  year={2025}
}

@article{luo2025deepcoder,
  title={Deepcoder: A fully open-source 14b coder at o3-mini level},
  author={Luo, Michael and Tan, Sijun and Huang, Roy and Patel, Ameen and Ariyak, Alpay and Wu, Qingyang and Shi, Xiaoxiang and Xin, Rachel and Cai, Colin and Weber, Maurice and others},
  journal={Notion Blog},
  year={2025}
}

@article{li2022competition,
  title={Competition-level code generation with alphacode},
  author={Li, Yujia and Choi, David and Chung, Junyoung and Kushman, Nate and Schrittwieser, Julian and Leblond, R{\'e}mi and Eccles, Tom and Keeling, James and Gimeno, Felix and Dal Lago, Agustin and others},
  journal={Science},
  volume={378},
  number={6624},
  pages={1092--1097},
  year={2022},
  publisher={American Association for the Advancement of Science}
}

@misc{penedo2025codeforces,
      title={CodeForces}, 
      author={Guilherme Penedo and Anton Lozhkov and Hynek Kydlíček and Loubna Ben Allal and Edward Beeching and Agustín Piqueres Lajarín and Quentin Gallouédec and Nathan Habib and Lewis Tunstall and Leandro von Werra},
      year={2025},
      publisher = {Hugging Face},
      journal = {Hugging Face repository},
      howpublished = {\url{https://huggingface.co/datasets/open-r1/codeforces}}
}

@article{minerva,
  title={Solving quantitative reasoning problems with language models},
  author={Lewkowycz, Aitor and Andreassen, Anders and Dohan, David and Dyer, Ethan and Michalewski, Henryk and Ramasesh, Vinay and Slone, Ambrose and Anil, Cem and Schlag, Imanol and Gutman-Solo, Theo and others},
  journal={Advances in neural information processing systems},
  volume={35},
  pages={3843--3857},
  year={2022}
}

@misc{maa2024aime,
  author       = {{MAA}},
  title        = {American Invitational Mathematics Examination ({AIME})},
  year         = {2024},
  howpublished = {\url{https://artofproblemsolving.com/wiki/index.php/AIME_Problems_and_Solutions}},
  note         = {Accessed: 2025}
}

@misc{maa2025aime,
  author       = {{MAA}},
  title        = {American Invitational Mathematics Examination ({AIME})},
  year         = {2025},
  howpublished = {\url{https://artofproblemsolving.com/wiki/index.php/AIME_Problems_and_Solutions}},
  note         = {Accessed: 2025}
}

@article{chen2021evaluating,
  title={Evaluating large language models trained on code},
  author={Chen, Mark},
  journal={arXiv preprint arXiv:2107.03374},
  year={2021}
}

@article{he2026far,
  title={How Far Can Unsupervised RLVR Scale LLM Training?},
  author={He, Bingxiang and Zuo, Yuxin and Liu, Zeyuan and Zhao, Shangziqi and Fu, Zixuan and Yang, Junlin and Qian, Cheng and Zhang, Kaiyan and Fan, Yuchen and Cui, Ganqu and others},
  journal={arXiv preprint arXiv:2603.08660},
  year={2026}
}

\newpage
\appendix
\onecolumn

\section{Related work}

\paragraph{Code Agents}
Code agents extend LLMs to perform multi-step software engineering tasks, interacting with codebases, tools, and external environments. SWE-bench~\citep{jimenezSWEbenchCanLanguage2024} evaluates agents on real GitHub issues requiring coordinated changes across multiple files, and has become a standard benchmark for measuring autonomous software engineering capabilities. SWE-agent~\citep{yangSWEagentAgentComputerInterfaces2024} demonstrates effective agent architectures for this task. Training agents for such tasks is challenging due to long trajectories, sparse rewards, and the difficulty of obtaining comprehensive test suites for real-world repositories. This setting motivates interest in RLIF methods that could provide training signal without complete ground-truth verification, though empirical investigation of such approaches for code agents remains limited.

\paragraph{Reinforcement Learning for Code Generation}
Training code LLMs with reinforcement learning has been explored through execution-based rewards, where models receive feedback based on whether generated code passes test cases. CodeRL~\citep{leCodeRLMasteringCode2022} trains a critic to predict functional correctness and uses this as reward signal. StepCoder~\citep{dou2024stepcoder} introduces curriculum learning with RL to decompose complex tasks into manageable subtasks. PPOCoder~\citep{shojaeeExecutionbasedCodeGeneration2023} applies proximal policy optimization with execution feedback. More recent work such as CodeRL+~\citep{jiang2025coderl+} integrates execution semantics alignment to provide denser learning signals beyond binary pass/fail rewards. These approaches provide effective training signal when high-quality test suites are available, but face limitations in settings where test coverage is incomplete or test cases are expensive to obtain.
\section{Additional Experimental Details}
\label{app:experimental_details}
\subsection{Experiment Setup}
\subsubsection{Dataset}
We adopt the training corpus from~\citep{archer}, which comprises 6,753 programming problems drawn from competitive programming datasets.
For details on the data cleaning and filtering pipeline (including test case preprocessing, difficulty filtering, deduplication, and contamination prevention), we refer readers to the original paper~\citep{archer}.

\subsubsection{Evaluation Metric}
\label{app:evaluation metric}

We evaluate model performance using three complementary metrics:

\textbf{pass@K} measures the probability that at least one of $K$ sampled solutions is correct. We use the unbiased estimator from~\citep{chen2021evaluating}:
\begin{equation}
\text{pass@}K = \mathbb{E}\left[1 - \frac{\binom{n-c}{K}}{\binom{n}{K}}\right]
\end{equation}
where $n$ is the total number of samples and $c$ is the number of correct samples. This metric captures the model's ceiling performance when given multiple attempts.

\textbf{avg@K} computes the average accuracy across $K$ independent samples, which is equivalent to pass@1:
\begin{equation}
\text{avg@}K = \frac{1}{K}\sum_{i=1}^{K} \mathbf{1}[y_i = y^*] = \text{pass@}1
\end{equation}
where $y_i$ is the prediction of the $i$-th sample and $y^*$ is the ground truth. This metric reflects the expected success rate of a single generation attempt.

\textbf{cons@K} (consensus@K) denotes majority-vote accuracy over $K$ samples:
\begin{equation}
\text{cons@}K = \mathbf{1}\left[\arg\max_{y} \sum_{i=1}^{K} \mathbf{1}[y_i = y] = y^*\right]
\end{equation}
where the prediction is determined by selecting the most frequently occurring answer among $K$ samples. For mathematical reasoning tasks like AIME, we extract the final numerical answer from each sample and apply majority voting.

\subsubsection{Evaluation Prompt}

\begin{tcolorbox}[title=LiveCodeBench Evaluation Prompt, colback=gray!5]
\ttfamily
You are an expert Python programmer. You will be given a question (problem specification) and will generate a correct Python program that matches the specification and passes all tests.

\textless question\textgreater

\#\#\# Format: Read the inputs from stdin solve the problem and write the answer to stdout (do not directly test on the sample inputs). Enclose your code within delimiters as follows. Ensure that when the python program runs, it reads the inputs, runs the algorithm and writes output to STDOUT.

\#\ YOUR CODE HERE

\#\#\# Answer: (use the provided format with backticks)
\end{tcolorbox}

\begin{tcolorbox}[title=AIME Evaluation Prompt, colback=gray!5]
\ttfamily
<question>\\[0.5em]
Please reason step by step, and put your final answer within \textbackslash boxed\{\}
\end{tcolorbox}

\subsection{Certainty-Correctness Analysis}
\label{app:certainty-correctness}

\begin{figure*}[htbp]
    \centering
    \includegraphics[width=\textwidth]{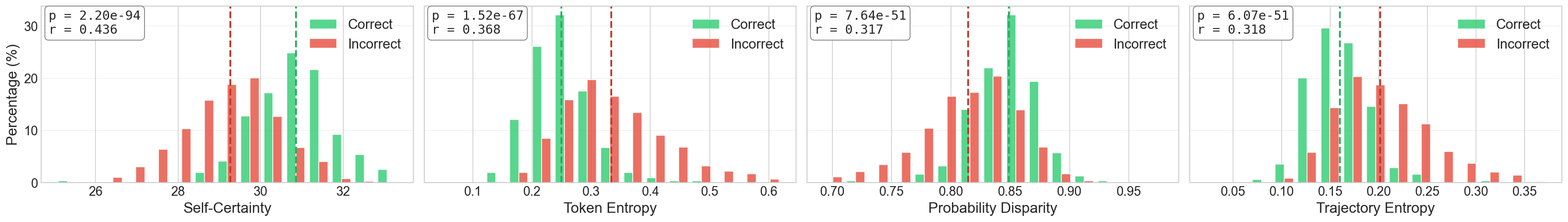}
    \caption{Distributions of four certainty metrics for correct (green) and incorrect (red) solutions on LiveCodeBench v5, using the Probability Disparity method at step 105. Dashed lines indicate class means; $r$ denotes rank-biserial correlation.}
    \label{fig:prob-disparity-certainty}
\end{figure*}

\begin{figure*}[htbp]
    \centering
    \includegraphics[width=\textwidth]{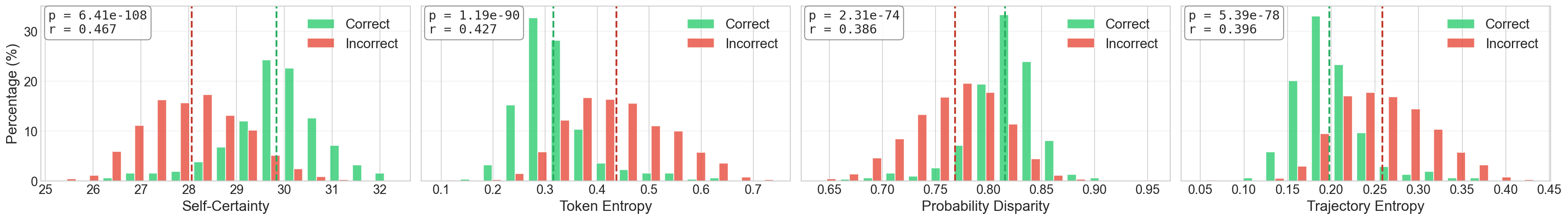}
    \caption{Distributions of four certainty metrics for correct (green) and incorrect (red) solutions on LiveCodeBench v5, using the Token Entropy method at step 105. Dashed lines indicate class means; $r$ denotes rank-biserial correlation.}
    \label{fig:token-entropy-certainty}
\end{figure*}

\begin{figure*}[htbp]
    \centering
    \includegraphics[width=\textwidth]{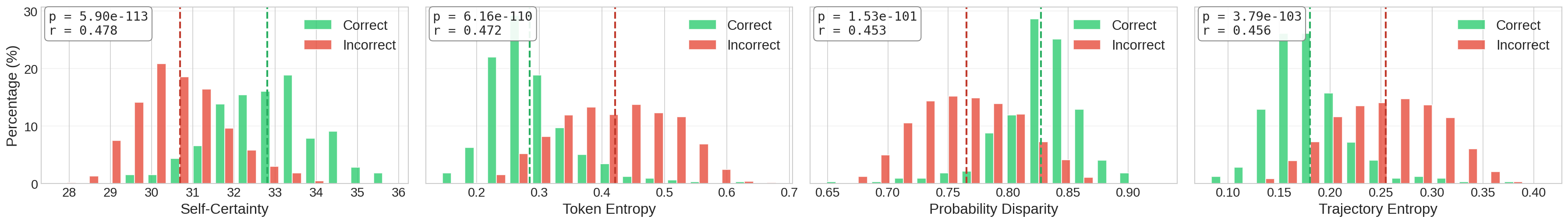}
    \caption{Distributions of four certainty metrics for correct (green) and incorrect (red) solutions on LiveCodeBench v5, using the Trajectory Entropy method at step 105. Dashed lines indicate class means; $r$ denotes rank-biserial correlation.}
    \label{fig:trajectory-entropy-certainty}
\end{figure*}

In this section, we provide additional certainty-correctness analyses for methods.
Figures~\ref{fig:prob-disparity-certainty}, \ref{fig:token-entropy-certainty}, \ref{fig:trajectory-entropy-certainty} present certainty-correctness analyses for the remaining RLIF methods. All three methods show the same phenomenon observed in the main text: the certainty distributions shift rightward for both correct and incorrect solutions.

Compared to the GRPO baseline, all RLIF methods show weaker certainty-correctness correlations. The distributions reveal that RLIF training raises certainty scores equally for correct and incorrect solutions, causing the two classes to overlap more heavily. This pattern directly reflects the entropy collapse shown in Section~\ref{sec:training_dynamics}: as the model's output distribution narrows, it becomes confident regardless of solution correctness.

\subsection{Resuming RLVR from RLIF Checkpoints}
\label{app:bootstrap}

\begin{figure*}[htbp]
    \centering
    \includegraphics[width=\textwidth]{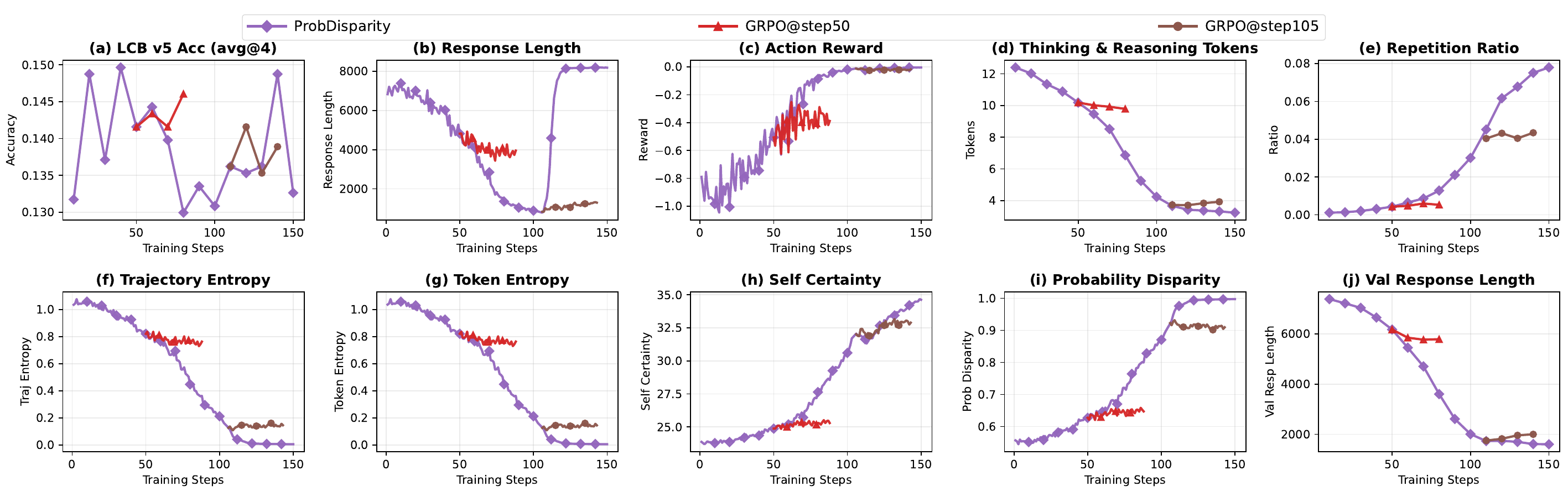}
    \caption{Bootstrapping GRPO from ProbDisparity checkpoints at steps 50 and 105. Metrics include (a) LiveCodeBench v5 avg@4 accuracy, (b) response length, (c) action reward, (d) thinking \& reasoning tokens, (e) repetition ratio, (f) trajectory entropy, (g) token entropy, (h) self-certainty, (i) probability disparity, and (j) validation response length. Each GRPO run continues for 40 steps from the corresponding checkpoint.}
    \label{fig:bootstrap_grpo_probdisparity}
\end{figure*}

\begin{figure*}[htbp]
    \centering
    \includegraphics[width=\textwidth]{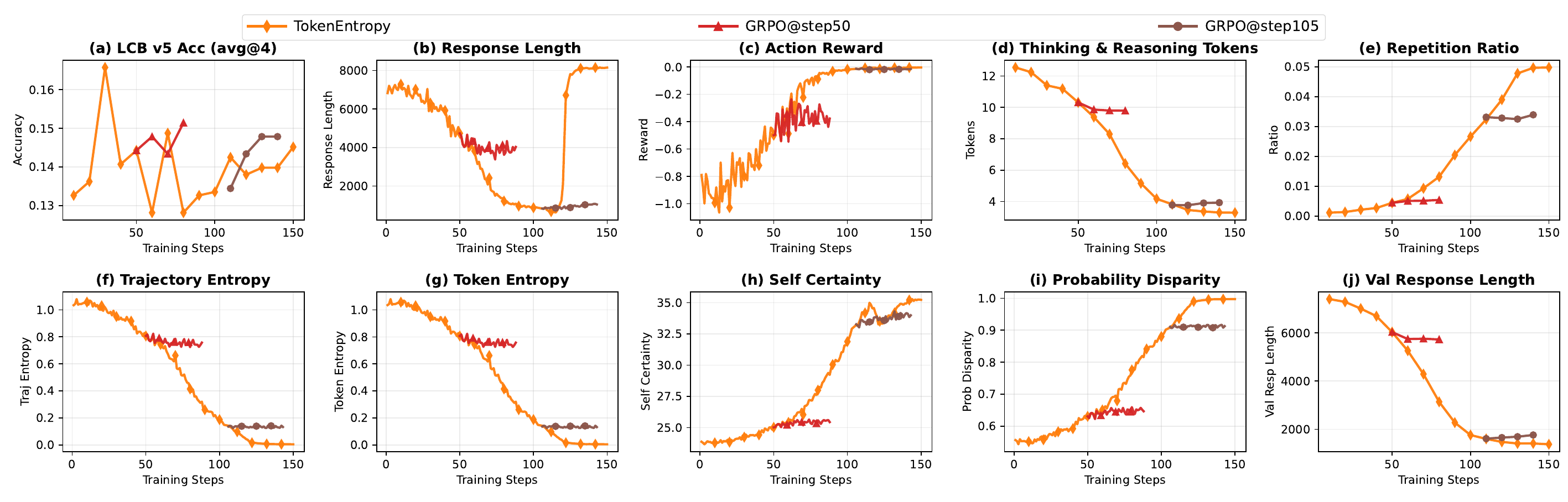}
    \caption{Bootstrapping GRPO from Token Entropy checkpoints at steps 50 and 105. Metrics include (a) LiveCodeBench v5 avg@4 accuracy, (b) response length, (c) action reward, (d) thinking \& reasoning tokens, (e) repetition ratio, (f) trajectory entropy, (g) token entropy, (h) self-certainty, (i) probability disparity, and (j) validation response length. Each GRPO run continues for 40 steps from the corresponding checkpoint.}
    \label{fig:bootstrap_grpo_tokenentropy}
\end{figure*}

\begin{figure*}[htbp]
    \centering
    \includegraphics[width=\textwidth]{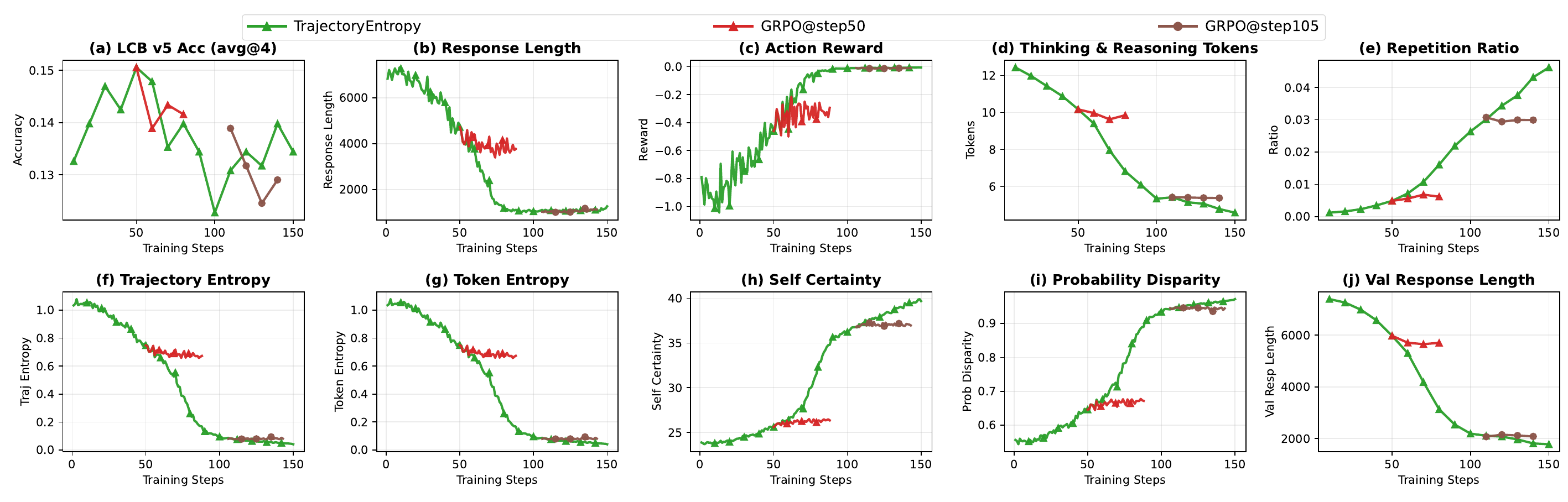}
    \caption{Bootstrapping GRPO from Trajectory Entropy checkpoints at steps 50 and 105. Metrics include (a) LiveCodeBench v5 avg@4 accuracy, (b) response length, (c) action reward, (d) thinking \& reasoning tokens, (e) repetition ratio, (f) trajectory entropy, (g) token entropy, (h) self-certainty, (i) probability disparity, and (j) validation response length. Each GRPO run continues for 40 steps from the corresponding checkpoint.}
    \label{fig:bootstrap_grpo_trajectoryentropy}
\end{figure*}

Figures~\ref{fig:bootstrap_grpo_probdisparity}, \ref{fig:bootstrap_grpo_tokenentropy}, and \ref{fig:bootstrap_grpo_trajectoryentropy} present training dynamics for bootstrapping GRPO from ProbDisparity, Token Entropy, and Trajectory Entropy checkpoints, respectively.

For Probability Disparity in Figure~\ref{fig:bootstrap_grpo_probdisparity}, GRPO@step50 peaks at 0.146 accuracy while GRPO@step105 reaches 0.142, a narrow gap. The step 50 checkpoint maintains trajectory entropy $\approx$0.76--0.80 and $\approx$10 reasoning tokens, supporting stable exploration. The step 105 checkpoint starts with severely collapsed entropy ($\approx$0.13) and $\approx$4 reasoning tokens, yet shows gradual recovery with slowly increasing response length and accuracy throughout training.

For Token Entropy in Figure~\ref{fig:bootstrap_grpo_tokenentropy}, the gap narrows further: GRPO@step50 reaches 0.151 and GRPO@step105 achieves 0.148 with a steady upward trend. Despite starting from collapsed entropy ($\approx$0.13) and suppressed reasoning tokens ($\approx$4), GRPO@step105 converges to near-equivalent performance.

For Trajectory Entropy in Figure~\ref{fig:bootstrap_grpo_trajectoryentropy}, the collapse at step 105 is most severe (trajectory entropy $\approx$0.08, self-certainty $\approx$37). GRPO@step50 peaks at 0.143, while GRPO@step105 briefly reaches 0.139 before declining to 0.129 by the end of training, the only case where accuracy trends downward. This suggests that when diversity collapse is sufficiently extreme, GRPO cannot recover meaningful exploration.

\subsection{Test-Time Training}
\label{app:test time train}
\begin{figure*}[htbp]
    \centering
    \includegraphics[width=\textwidth]{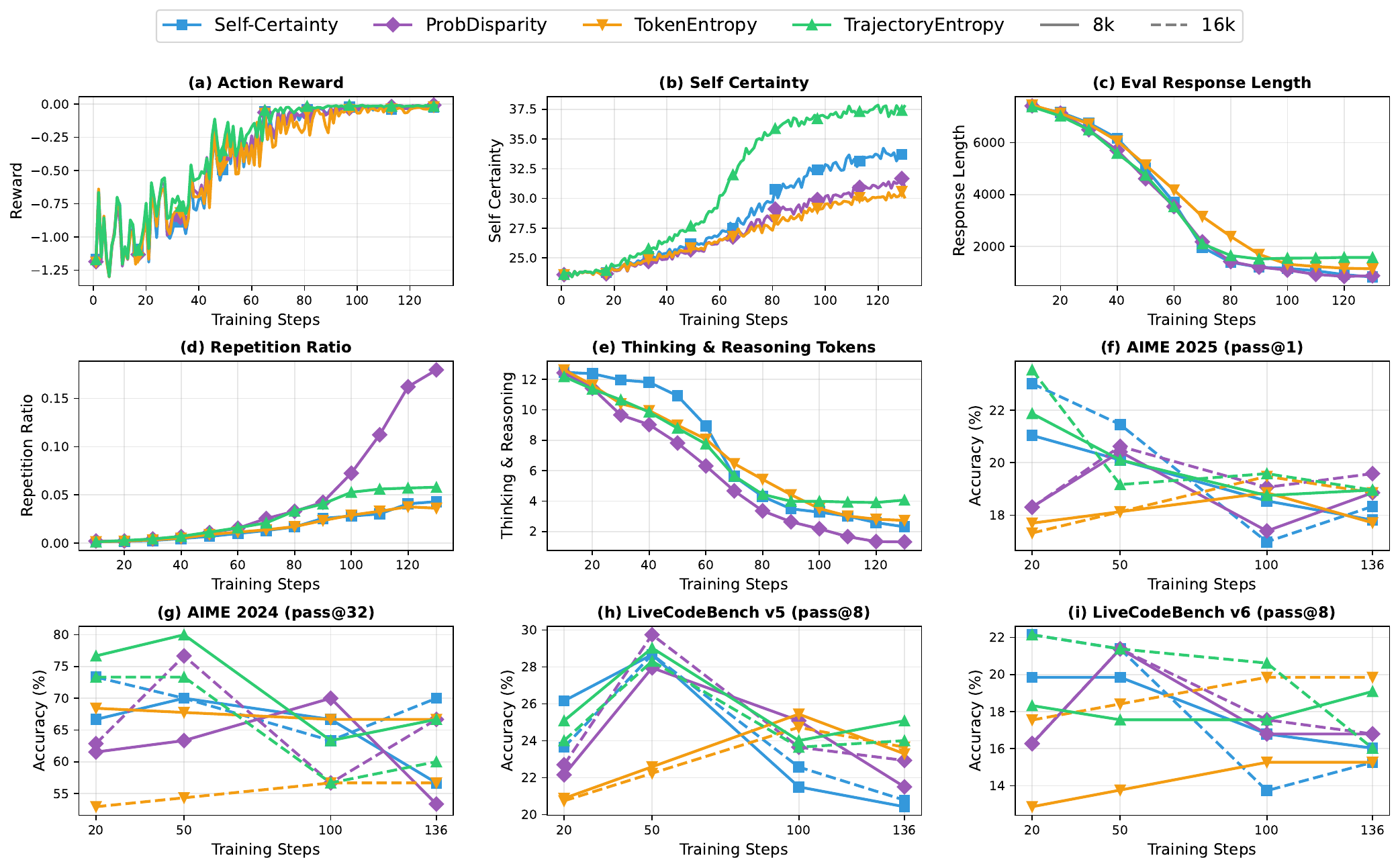}
    \caption{Additional metrics for test-time training on LiveCodeBench v5. Top row: action reward, self-certainty, response length, repetition ratio, and reasoning token count. Bottom row: evaluation performance on AIME 2025 (pass@1), AIME 2024 (pass@32), LCB v5 (pass@8), and LCB v6 (pass@8). Solid lines denote 8k context; dashed lines denote 16k context.}
    \label{fig:lcb-appendix}
\end{figure*}

Figure~\ref{fig:lcb-appendix} provides additional metrics for Section~\ref{exp:Test-Time Training on LCB}. Response length collapse occurs around step 70 for certainty-based methods. Notably, Probability Disparity develops severe repetition (ratio reaching 0.18 vs.\ under 0.05 for others), indicating reward hacking after initial training.

The count of thinking and reasoning tokens (e) reveals that all certainty-based methods show declining reasoning tokens as training progresses, suggesting that optimizing for certainty may discourage extended deliberation. At pass@32 on AIME 2024 (g), Trajectory-Entropy achieves strong early performance ($\sim$80\% at step 50) before declining. The 16k context advantage is most pronounced in early training but largely disappears by step 136. On LiveCodeBench at pass@8 (h--i), all certainty methods show degradation as training progresses, particularly on the out-of-domain LCB v6 benchmark.

\subsection{Ablation Studies}
\label{app:ablation}

We conduct ablation studies on rollout count $N \in \{8, 12, 16\}$, sampling temperature $\tau \in \{0.8, 1.0, 1.2\}$, KL coefficient $\beta_{\text{KL}} \in \{0, 0.005\}$, and PPO epochs $E \in \{1, 3\}$ for all methods. Each parameter is varied independently with others fixed at default values ($N=16$, $\tau=1.0$, $\beta_{\text{KL}}=0$, $E=1$). All experiments are conducted on LiveCodeBench v5 with accuracy evaluated using avg@4.

\subsubsection{Self-Certainty}
\label{app:ablation_certainty}

\begin{figure*}[htbp]
    \centering
    \includegraphics[width=\textwidth]{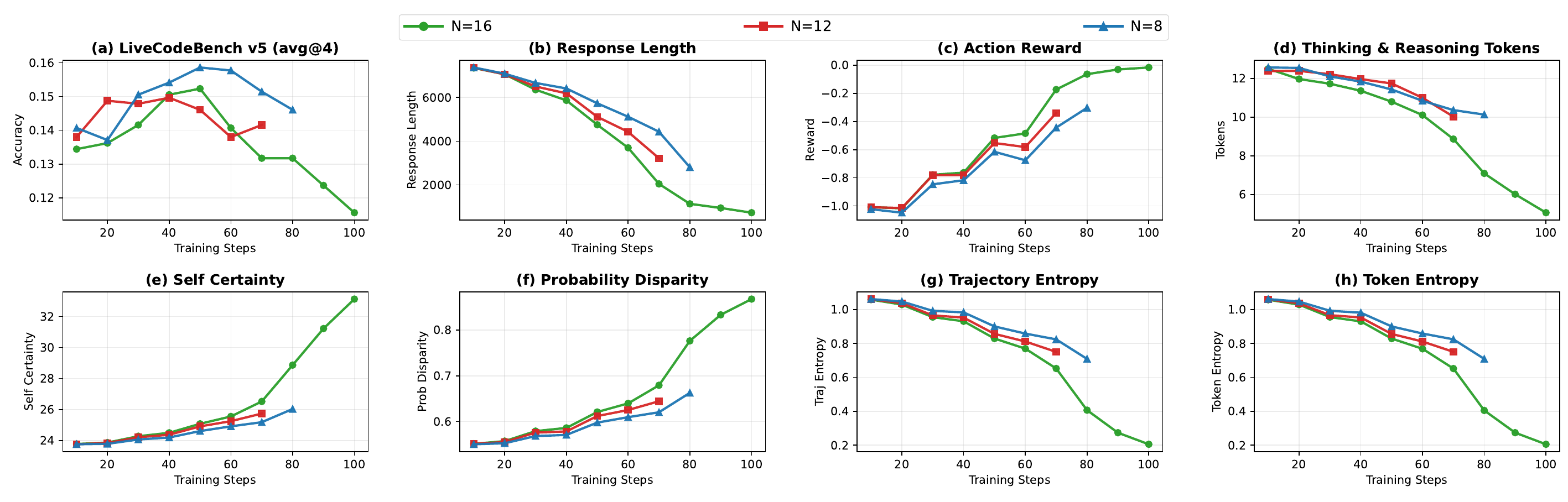}
    \caption{Ablation study on the number of rollouts $N \in \{8, 12, 16\}$ for Self-Certainty. Subplots show (a) LiveCodeBench v5 avg@4 accuracy, (b) response length, (c) action reward, (d) thinking \& reasoning token, (e) self-certainty, (f) probability disparity, (g) trajectory entropy, (h) token entropy.}
    \label{fig:ablation-intuitor-num}
\end{figure*}

Figure~\ref{fig:ablation-intuitor-num} shows the effect of rollout count on Self-Certainty training. With $N=16$, accuracy peaks at 15.2\% around step 50 before collapsing to 9.3\% by step 200, accompanied by response length dropping from approximately 7{,}350 to under 650 tokens. The $N=12$ and $N=8$ configurations (trained for 70 and 80 steps respectively) maintain more stable accuracy, finishing at 14.2\% and 14.6\%, with response lengths degrading more gradually to around 3{,}000--3{,}200 tokens. The self-certainty metric rises steeply for all configurations, but the rate is fastest for $N=16$, consistent with more severe collapse.

\begin{figure*}[htbp]
    \centering
    \includegraphics[width=\textwidth]{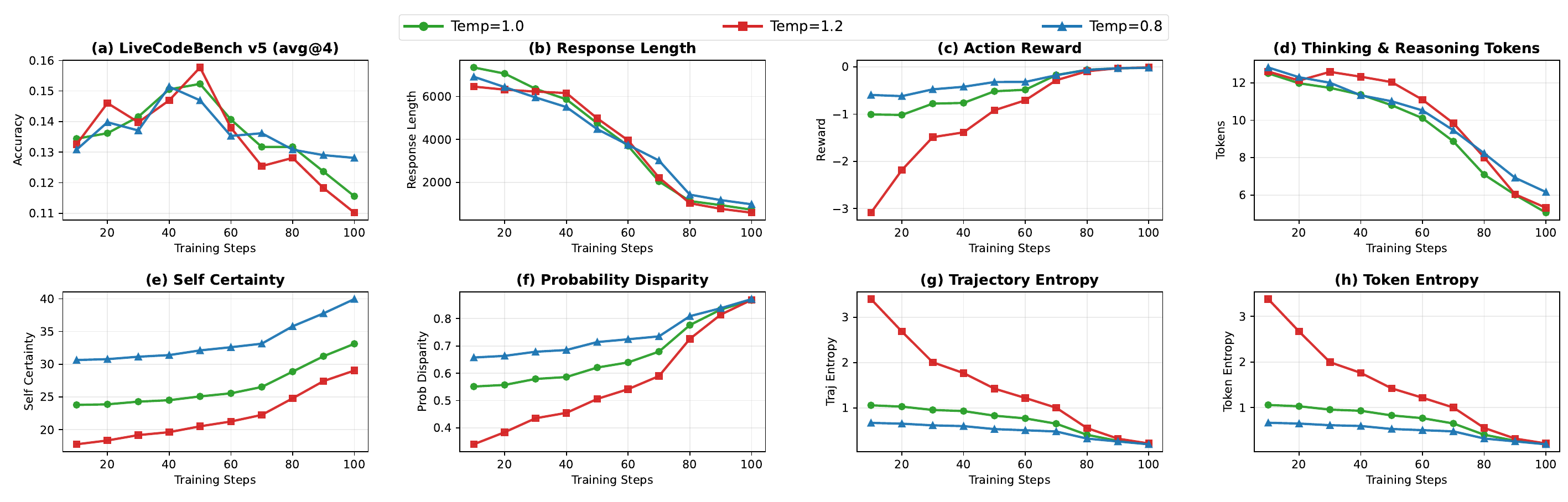}
    \caption{Ablation study on sampling temperature $\tau \in \{0.8, 1.0, 1.2\}$ for Self-Certainty. Subplots show (a) LiveCodeBench v5 avg@4 accuracy, (b) response length, (c) action reward, (d) thinking \& reasoning token, (e) self-certainty, (f) probability disparity, (g) trajectory entropy, (h) token entropy.}
    \label{fig:ablation-intuitor-temp}
\end{figure*}

Figure~\ref{fig:ablation-intuitor-temp} examines temperature effects. Higher temperature ($\tau=1.2$) reaches a slightly higher peak accuracy (15.8\% at step 50) but ultimately collapses more severely, finishing at 11.0\% with response length under 600 tokens. Lower temperature ($\tau=0.8$) peaks at 15.1\% and degrades to 12.8\% by step 100 with responses around 985 tokens. All configurations converge to similar degraded performance, indicating that increased exploration does not address the fundamental reward hacking issue.

\begin{figure*}[htbp]
    \centering
    \includegraphics[width=\textwidth]{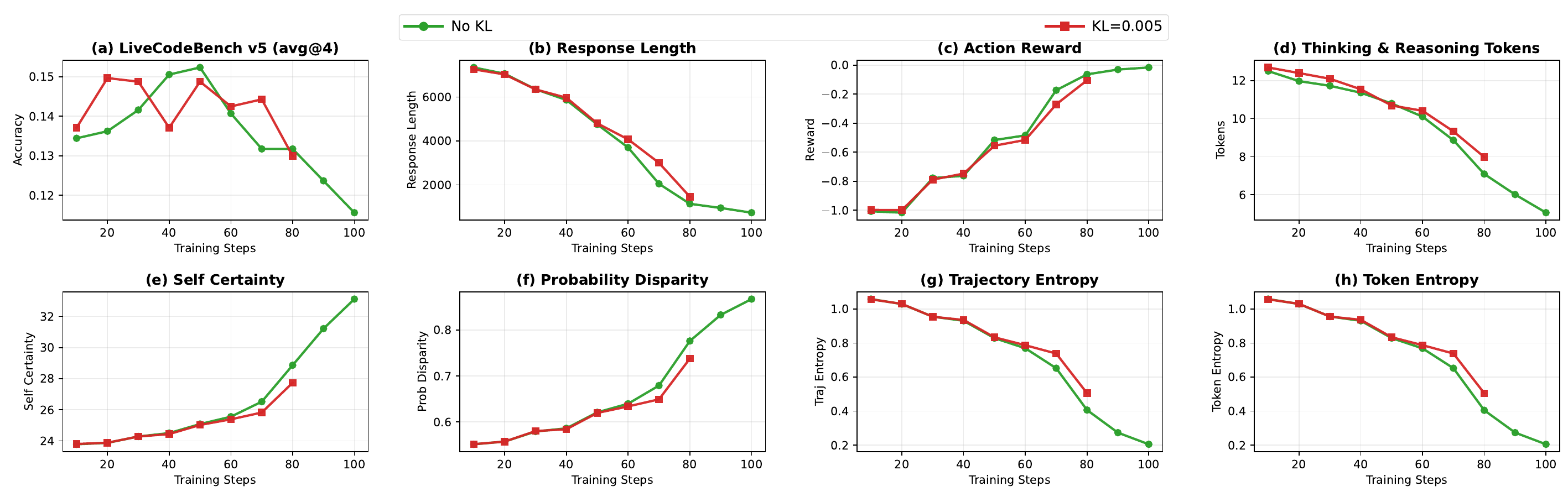}
    \caption{Ablation study on KL coefficient $\beta_{\text{KL}} \in \{0, 0.005\}$ for Self-Certainty. Subplots show (a) LiveCodeBench v5 avg@4 accuracy, (b) response length, (c) action reward, (d) thinking \& reasoning token, (e) self-certainty, (f) probability disparity, (g) trajectory entropy, (h) token entropy.}
    \label{fig:ablation-intuitor-kl}
\end{figure*}

Figure~\ref{fig:ablation-intuitor-kl} shows that applying a KL penalty ($\beta_{\text{KL}}=0.005$) moderately mitigates collapse. The KL-regularized run peaks at 15.0\% and retains 13.0\% accuracy at step 80 with response lengths around 1{,}460 tokens, compared to 9.3\% and 607 tokens for the baseline at step 200. The KL constraint slows self-certainty growth and entropy reduction, but does not fully prevent degradation.

\begin{figure*}[htbp]
    \centering
    \includegraphics[width=\textwidth]{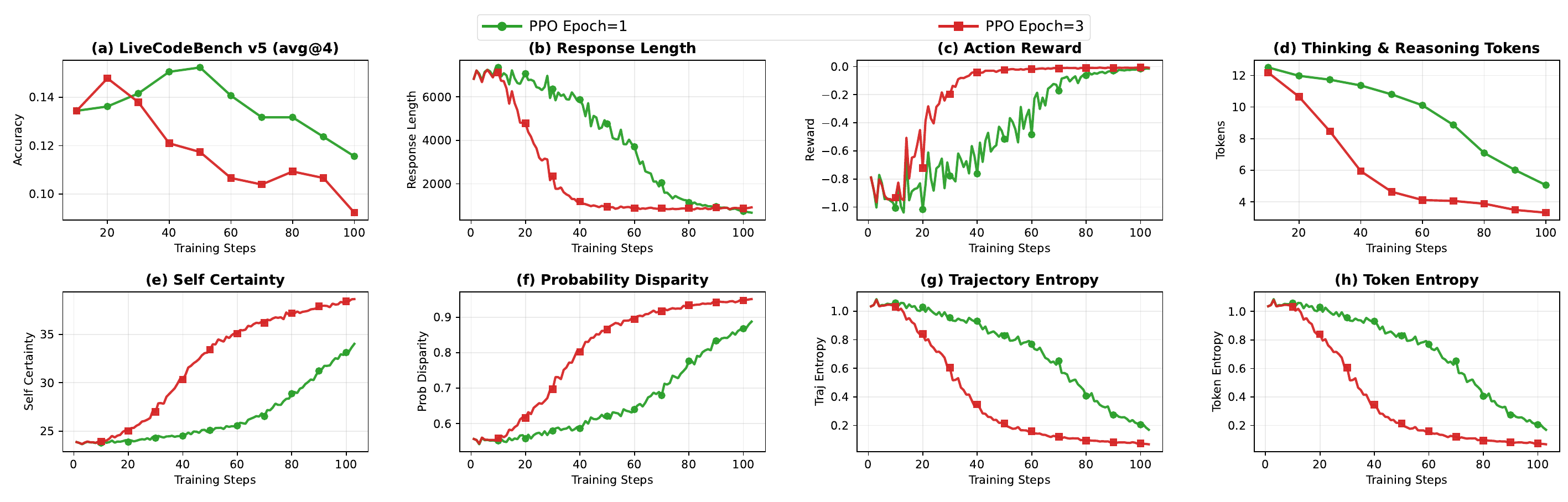}
    \caption{Ablation study on PPO epochs $E \in \{1, 3\}$ for Self-Certainty. Subplots show (a) LiveCodeBench v5 avg@4 accuracy, (b) response length, (c) action reward, (d) thinking \& reasoning token, (e) self-certainty, (f) probability disparity, (g) trajectory entropy, (h) token entropy.}
    \label{fig:ablation-intuitor-ppo}
\end{figure*}

Figure~\ref{fig:ablation-intuitor-ppo} shows that increasing to $E=3$ PPO epochs does not help Self-Certainty. The $E=3$ run peaks at 14.8\% at step 20 and collapses to 9.2\% by step 103, matching the final performance of $E=1$ but reaching degradation faster. Both configurations show near-zero trajectory entropy at the end of training.

\subsubsection{Token-Level Entropy}
\label{app:ablation_token}

\begin{figure*}[htbp]
    \centering
    \includegraphics[width=\textwidth]{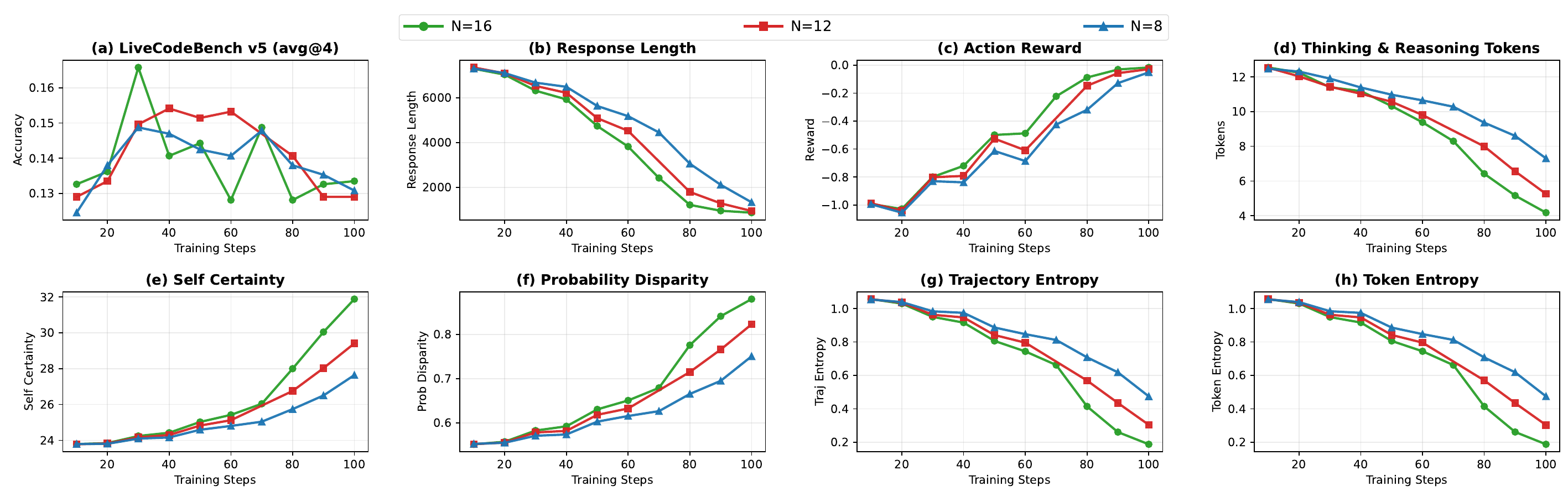}
    \caption{Ablation study on the number of rollouts $N \in \{8, 12, 16\}$ for Token-Level Entropy. Subplots show (a) LiveCodeBench v5 avg@4 accuracy, (b) response length, (c) action reward, (d) thinking \& reasoning token, (e) self-certainty, (f) probability disparity, (g) trajectory entropy, (h) token entropy.}
    \label{fig:ablation-token-num}
\end{figure*}

Figure~\ref{fig:ablation-token-num} reveals that Token-Level Entropy exhibits a distinct failure mode depending on rollout count. With $N=16$, the model collapses into a max-length degenerate state: accuracy peaks at 16.6\% around step 30 and then gradually settles at 13.9\% by step 160, while response length increases from 7{,}290 to approximately 8{,}100 tokens with trajectory entropy approaching zero. In contrast, $N=12$ and $N=8$ exhibit a short-response collapse: accuracy peaks around 15.4\% and 14.9\% before declining to 12.9\% and 13.1\%, while response lengths fall sharply to under 1{,}400 tokens. The contrasting collapse directions suggest that reward hacking manifests differently depending on gradient signal strength.

\begin{figure*}[htbp]
    \centering
    \includegraphics[width=\textwidth]{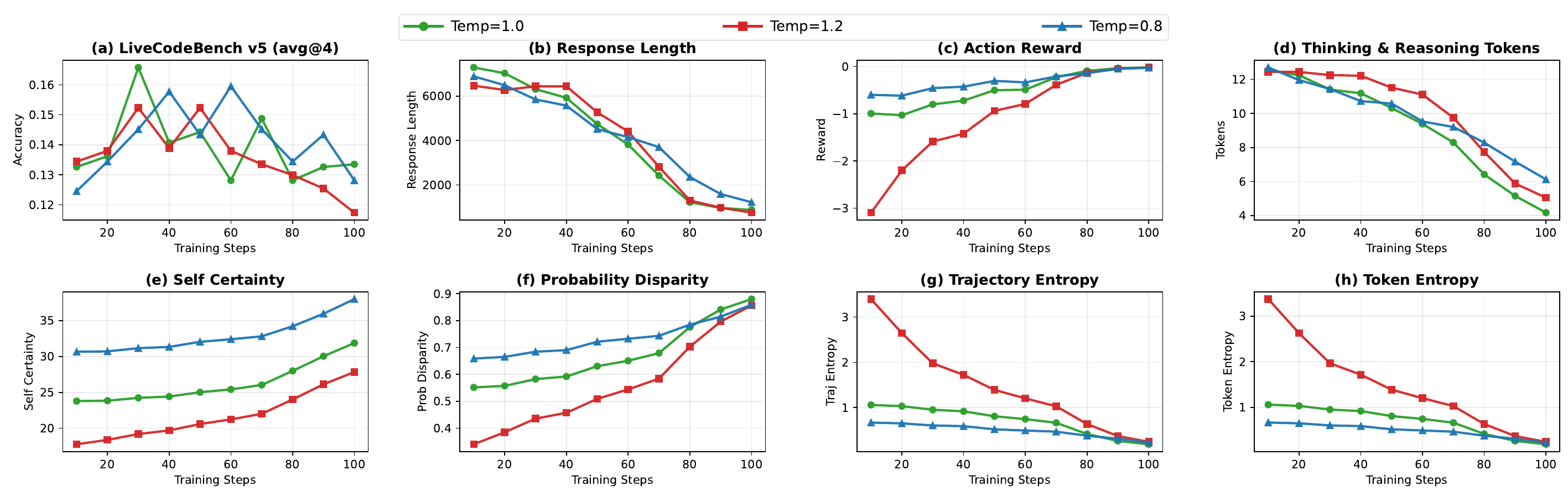}
    \caption{Ablation study on sampling temperature $\tau \in \{0.8, 1.0, 1.2\}$ for Token-Level Entropy. Subplots show (a) LiveCodeBench v5 avg@4 accuracy, (b) response length, (c) action reward, (d) thinking \& reasoning token, (e) self-certainty, (f) probability disparity, (g) trajectory entropy, (h) token entropy.}
    \label{fig:ablation-token-temp}
\end{figure*}

Figure~\ref{fig:ablation-token-temp} shows that all temperature variants collapse to short responses rather than max-length outputs. $\tau=0.8$ achieves the highest peak at 15.9\% (step 60) and finishes at 12.8\%, while $\tau=1.2$ peaks at 15.2\% (step 30) and collapses more severely to 11.7\% with response lengths under 750 tokens. All three configurations converge to comparable final accuracy (11.7--12.8\%) and severely shortened responses.

\begin{figure*}[htbp]
    \centering
    \includegraphics[width=\textwidth]{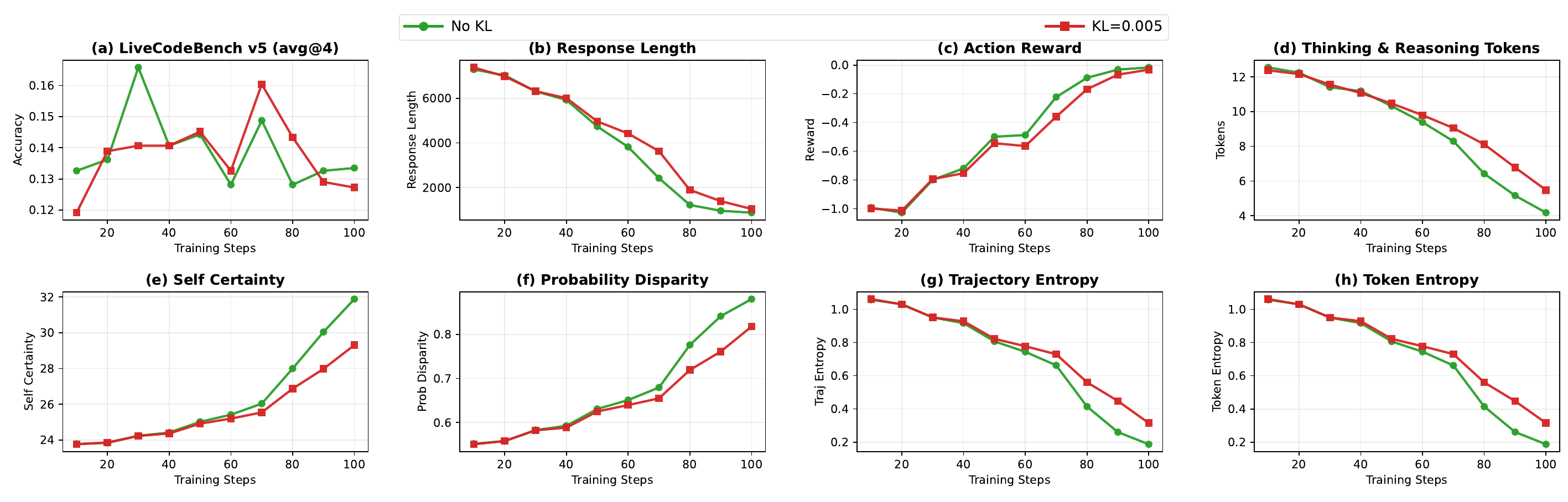}
    \caption{Ablation study on KL coefficient $\beta_{\text{KL}} \in \{0, 0.005\}$ for Token-Level Entropy. Subplots show (a) LiveCodeBench v5 avg@4 accuracy, (b) response length, (c) action reward, (d) thinking \& reasoning token, (e) self-certainty, (f) probability disparity, (g) trajectory entropy, (h) token entropy.}
    \label{fig:ablation-token-kl}
\end{figure*}

Figure~\ref{fig:ablation-token-kl} shows that KL regularization does not prevent collapse for Token-Level Entropy. The $\beta_{\text{KL}}=0.005$ run peaks at 16.0\% (step 70) and finishes at 12.7\% with response lengths around 1{,}038 tokens, similar in final accuracy to the temperature variants but without the max-length degenerate behavior of the $N=16$ baseline. The KL penalty effectively converts the collapse from max-length to short-response form without improving final accuracy.

\begin{figure*}[htbp]
    \centering
    \includegraphics[width=\textwidth]{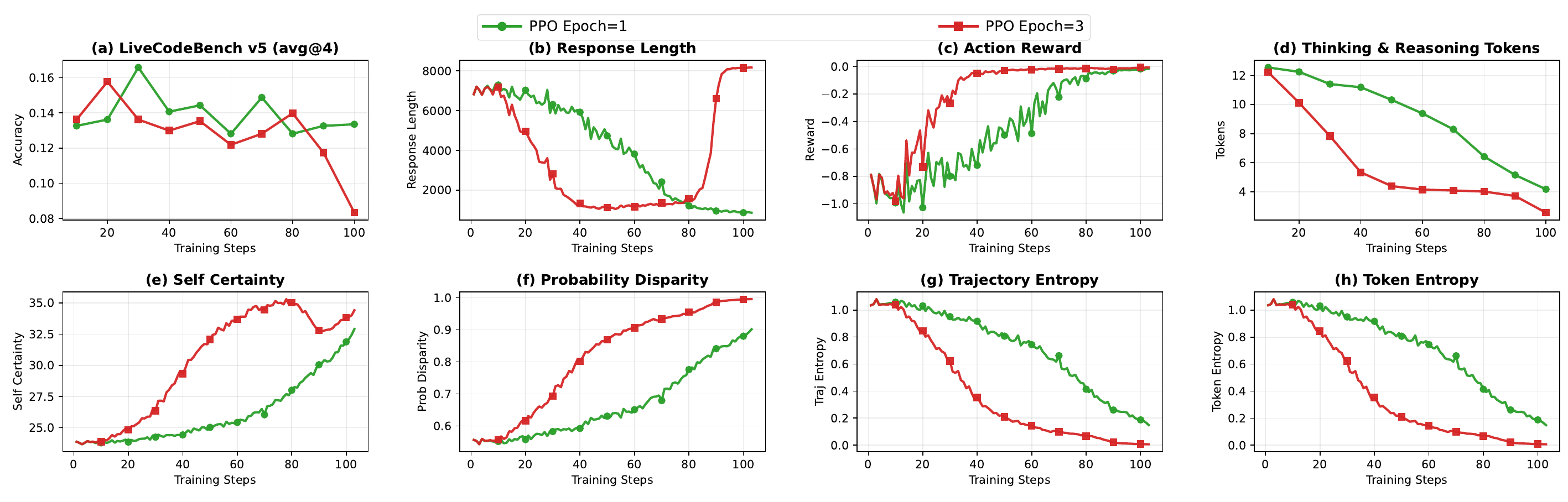}
    \caption{Ablation study on PPO epochs $E \in \{1, 3\}$ for Token-Level Entropy. Subplots show (a) LiveCodeBench v5 avg@4 accuracy, (b) response length, (c) action reward, (d) thinking \& reasoning token, (e) self-certainty, (f) probability disparity, (g) trajectory entropy, (h) token entropy.}
    \label{fig:ablation-token-ppo}
\end{figure*}

Figure~\ref{fig:ablation-token-ppo} shows that $E=3$ substantially worsens Token-Level Entropy. The run peaks early at 15.8\% (step 20) and then collapses severely into max-length degenerate outputs (response length $\approx$8{,}174 tokens, trajectory entropy $\approx$0.007), with final accuracy falling to 8.3\%, the worst final performance observed across all Token-Level Entropy configurations. This indicates that additional PPO updates per rollout dramatically accelerate the max-length collapse for this method.

\subsubsection{Trajectory-Level Entropy}
\label{app:ablation_traj}

\begin{figure*}[htbp]
    \centering
    \includegraphics[width=\textwidth]{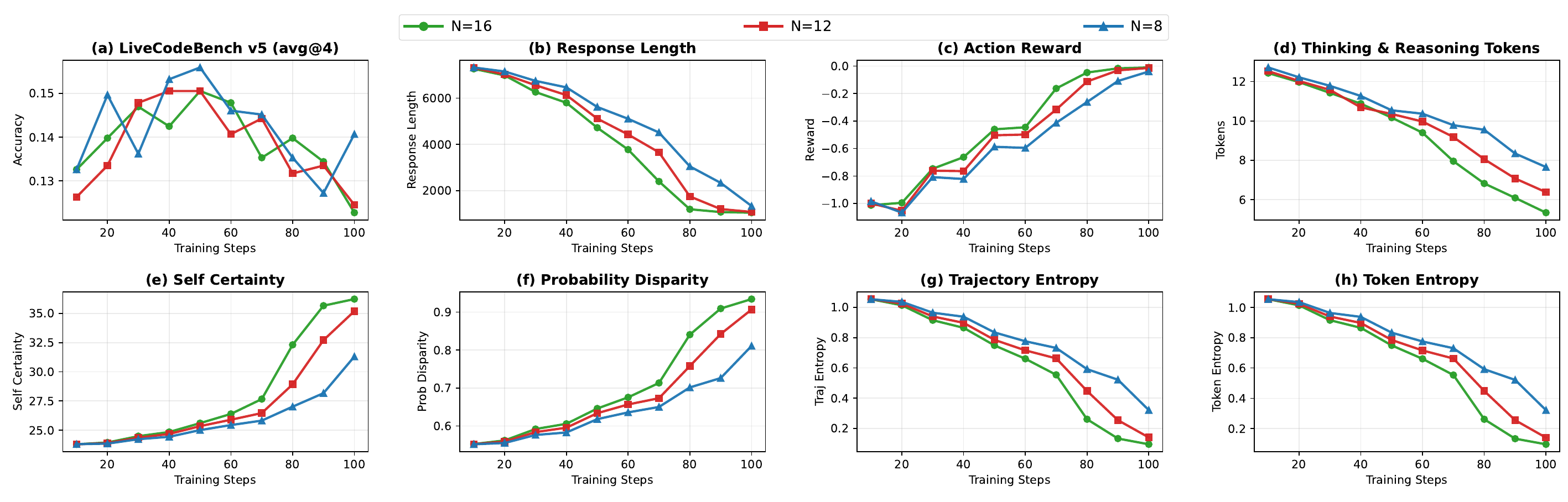}
    \caption{Ablation study on the number of rollouts $N \in \{8, 12, 16\}$ for Trajectory-Level Entropy. Subplots show (a) LiveCodeBench v5 avg@4 accuracy, (b) response length, (c) action reward, (d) thinking \& reasoning token, (e) self-certainty, (f) probability disparity, (g) trajectory entropy, (h) token entropy.}
    \label{fig:ablation-traj-num}
\end{figure*}

Figure~\ref{fig:ablation-traj-num} demonstrates that Trajectory-Level Entropy follows the same split-collapse pattern as Token-Level Entropy. With $N=16$, the model collapses into max-length degenerate outputs: accuracy peaks at 15.1\% around step 50 and stabilizes at 13.9\% by step 170, while response length increases to approximately 8{,}000 tokens and trajectory entropy drops to near zero. The $N=12$ and $N=8$ configurations instead show short-response collapse: accuracy peaks near 15.1\% and 15.6\% before declining to 12.5\% and 14.1\%, with response lengths falling to 1{,}080 and 1{,}340 tokens. The thinking and reasoning token count also decreases substantially from 12 to approximately 5--7, indicating significant loss of reasoning capability.

\begin{figure*}[htbp]
    \centering
    \includegraphics[width=\textwidth]{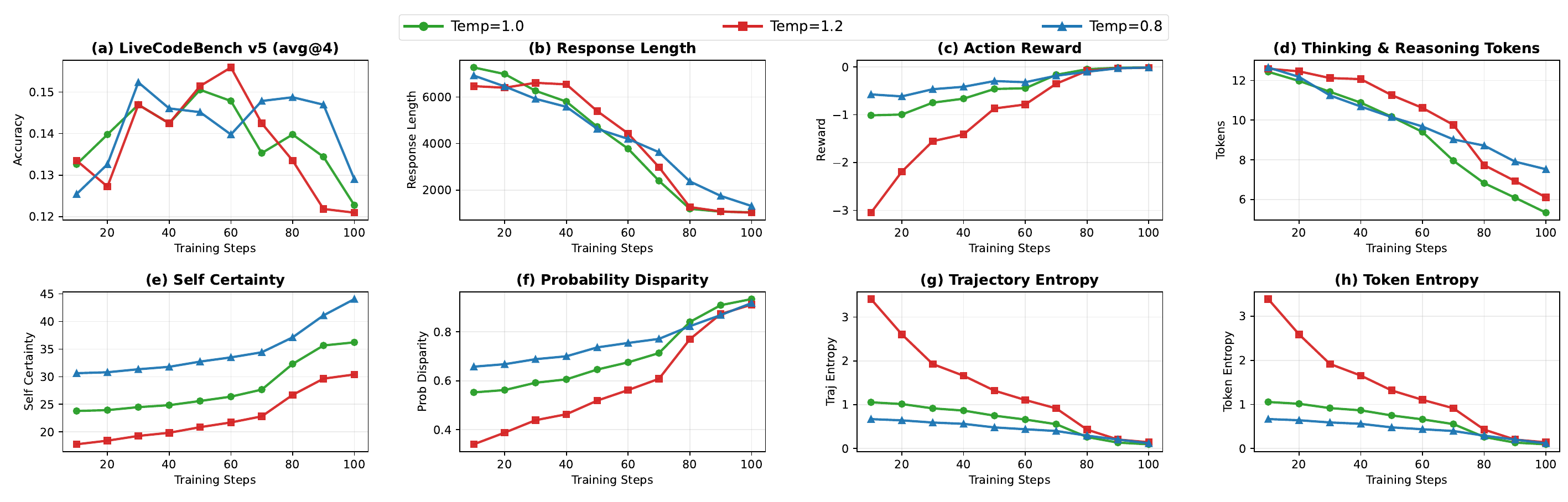}
    \caption{Ablation study on sampling temperature $\tau \in \{0.8, 1.0, 1.2\}$ for Trajectory-Level Entropy. Subplots show (a) LiveCodeBench v5 avg@4 accuracy, (b) response length, (c) action reward, (d) thinking \& reasoning token, (e) self-certainty, (f) probability disparity, (g) trajectory entropy, (h) token entropy.}
    \label{fig:ablation-traj-temp}
\end{figure*}

Figure~\ref{fig:ablation-traj-temp} indicates that temperature has limited ability to prevent collapse, and all three settings collapse to short responses. $\tau=0.8$ peaks at 15.2\% and ends at 12.9\%, while $\tau=1.2$ peaks at 15.6\% and declines to 12.1\%, slightly worse final accuracy despite a higher peak. All temperature configurations converge to response lengths around 1{,}000--1{,}300 tokens with comparable trajectory entropy ($\approx$0.12--0.14).

\begin{figure*}[htbp]
    \centering
    \includegraphics[width=\textwidth]{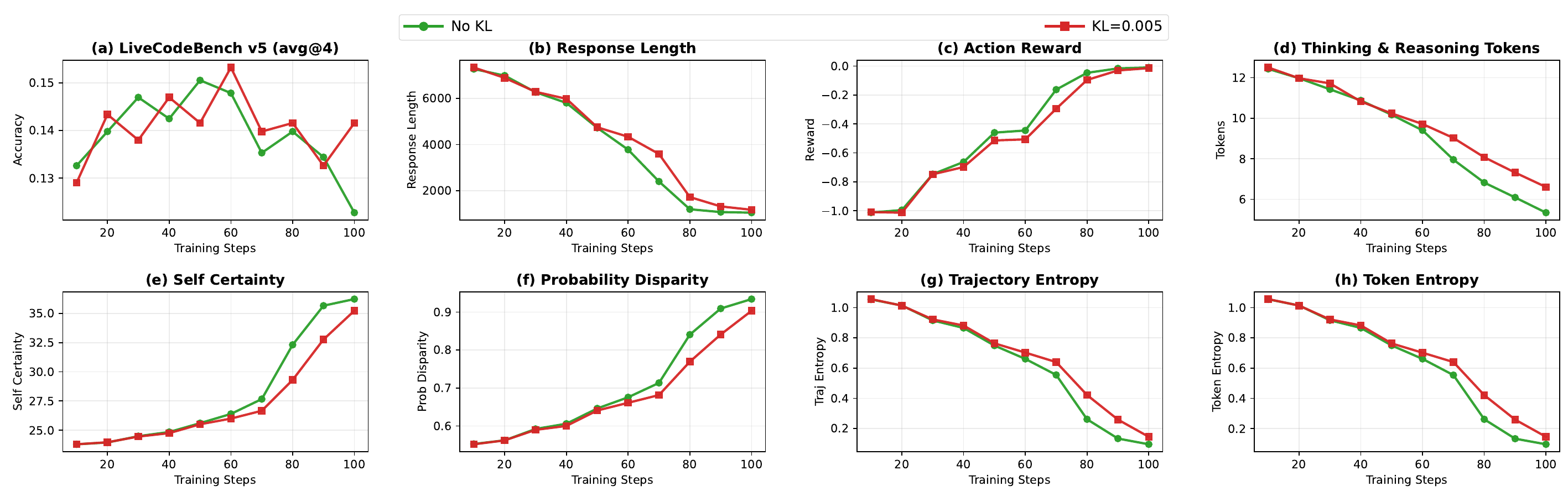}
    \caption{Ablation study on KL coefficient $\beta_{\text{KL}} \in \{0, 0.005\}$ for Trajectory-Level Entropy. Subplots show (a) LiveCodeBench v5 avg@4 accuracy, (b) response length, (c) action reward, (d) thinking \& reasoning token, (e) self-certainty, (f) probability disparity, (g) trajectory entropy, (h) token entropy.}
    \label{fig:ablation-traj-kl}
\end{figure*}

Figure~\ref{fig:ablation-traj-kl} reveals that KL regularization is beneficial for Trajectory-Level Entropy. The $\beta_{\text{KL}}=0.005$ run peaks at 15.3\% (step 60) and achieves the best final accuracy among all Trajectory-Level Entropy variants at 14.2\%, compared to 13.9\% for the $N=16$ baseline. The KL constraint prevents the max-length degenerate collapse, maintaining responses at approximately 1{,}170 tokens while preserving higher trajectory entropy (0.145 vs.\ 0.006 for the baseline).

\begin{figure*}[htbp]
    \centering
    \includegraphics[width=\textwidth]{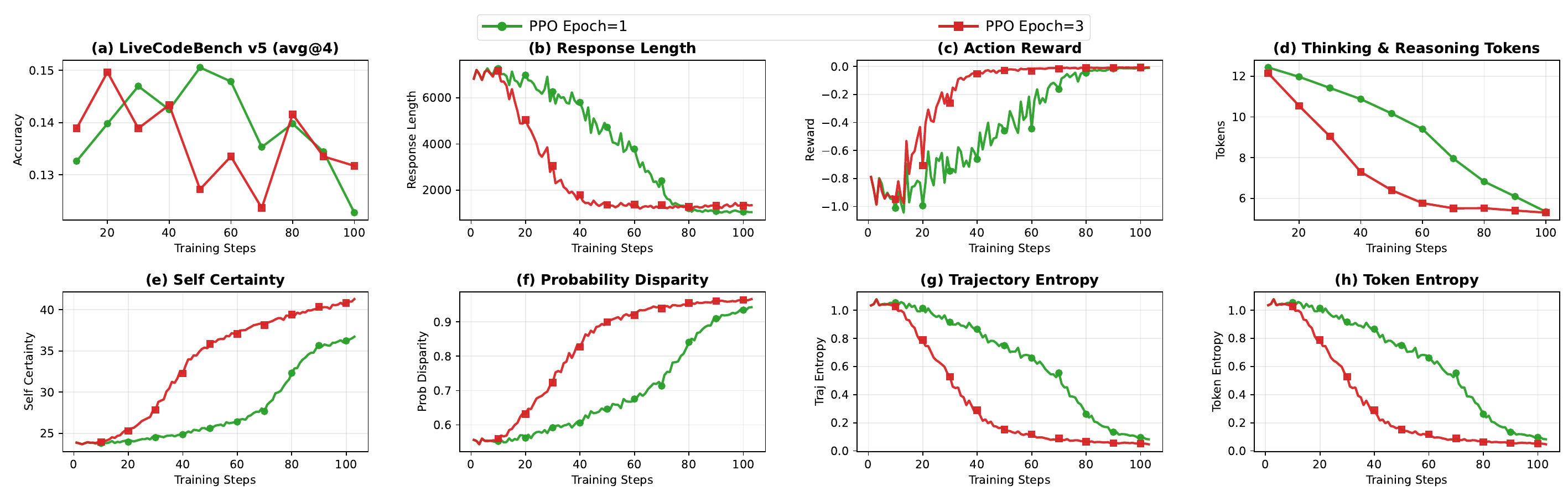}
    \caption{Ablation study on PPO epochs $E \in \{1, 3\}$ for Trajectory-Level Entropy. Subplots show (a) LiveCodeBench v5 avg@4 accuracy, (b) response length, (c) action reward, (d) thinking \& reasoning token, (e) self-certainty, (f) probability disparity, (g) trajectory entropy, (h) token entropy.}
    \label{fig:ablation-traj-ppo}
\end{figure*}

Figure~\ref{fig:ablation-traj-ppo} shows that $E=3$ causes Trajectory-Level Entropy to collapse to short responses (1{,}340 tokens final) rather than max-length outputs, with accuracy peaking at 15.0\% (step 20) and declining to 13.2\% by step 103. The trajectory entropy at the end (0.047) is higher than the $N=16$ baseline but lower than the $\beta_{\text{KL}}=0.005$ variant, indicating partial collapse.

\subsubsection{Probability Disparity}
\label{app:ablation_prob}

\begin{figure*}[htbp]
    \centering
    \includegraphics[width=\textwidth]{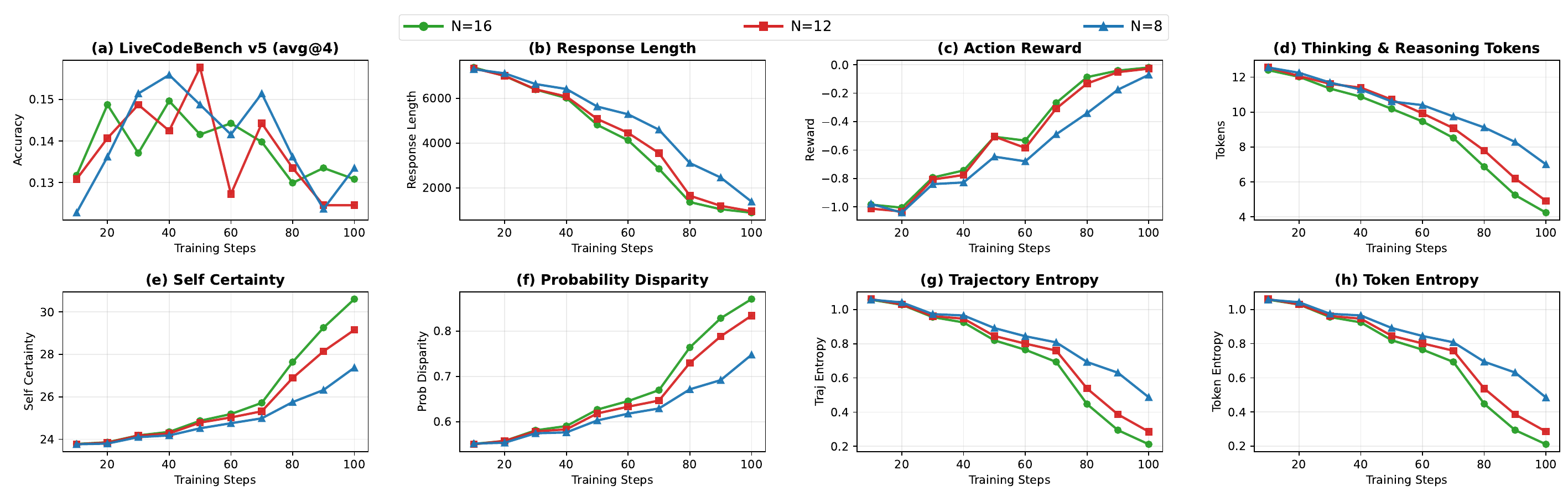}
    \caption{Ablation study on the number of rollouts $N \in \{8, 12, 16\}$ for Probability Disparity. Subplots show (a) LiveCodeBench v5 avg@4 accuracy, (b) response length, (c) action reward, (d) thinking \& reasoning token, (e) self-certainty, (f) probability disparity, (g) trajectory entropy, (h) token entropy.}
    \label{fig:ablation-prob-num}
\end{figure*}

Figure~\ref{fig:ablation-prob-num} shows that Probability Disparity exhibits the same bifurcated collapse pattern. With $N=16$, the model collapses into max-length degenerate outputs: accuracy peaks at 15.0\% around step 40 and stabilizes at 13.3\% by step 150 while response length grows to approximately 8{,}180 tokens with near-zero trajectory entropy (0.005). The $N=12$ and $N=8$ configurations instead collapse to short responses (950 and 1{,}380 tokens), with accuracy peaking at 15.8\% and 15.6\% before declining to 12.5\% and 13.4\%. Among the entropy-based methods, the $N=16$ max-length degenerate state maintains somewhat better final accuracy than the short-response collapse of $N=12$.

\begin{figure*}[htbp]
    \centering
    \includegraphics[width=\textwidth]{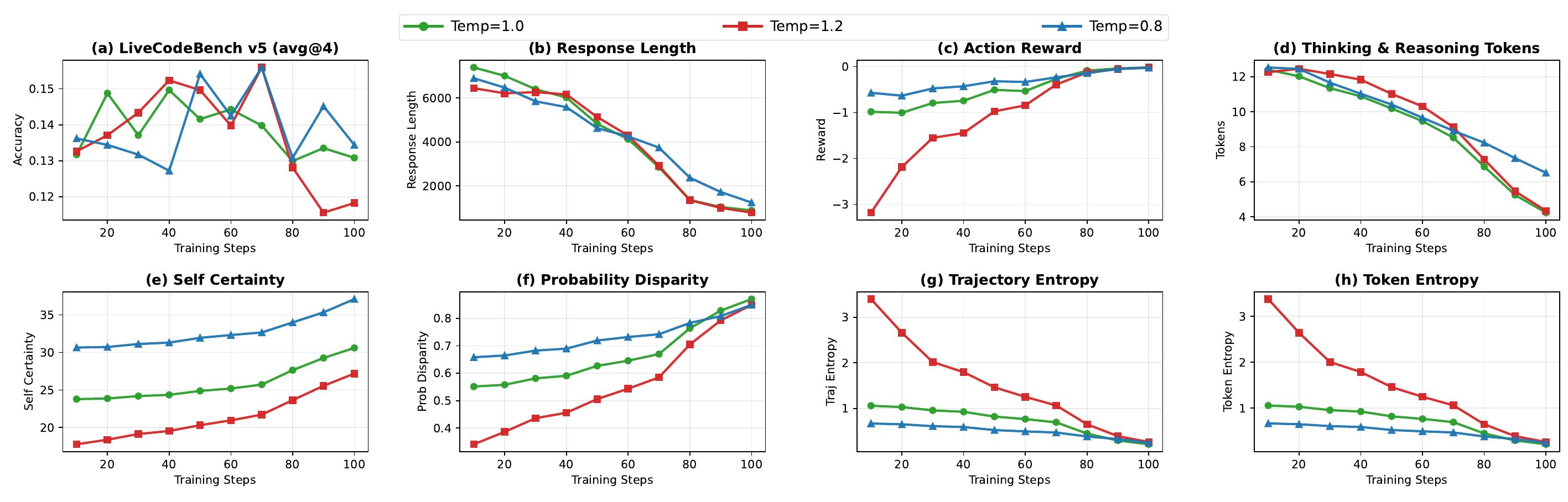}
    \caption{Ablation study on sampling temperature $\tau \in \{0.8, 1.0, 1.2\}$ for Probability Disparity. Subplots show (a) LiveCodeBench v5 avg@4 accuracy, (b) response length, (c) action reward, (d) thinking \& reasoning token, (e) self-certainty, (f) probability disparity, (g) trajectory entropy, (h) token entropy.}
    \label{fig:ablation-prob-temp}
\end{figure*}

Figure~\ref{fig:ablation-prob-temp} reveals that all temperature variants collapse to short responses with similar final accuracy. $\tau=0.8$ peaks at 15.6\% (step 70) and ends at 13.4\%, while $\tau=1.2$ also peaks at 15.6\% (step 70) but collapses more severely to 11.8\% with response lengths under 800 tokens. The inverted relationship between temperature and final accuracy confirms that increased sampling diversity does not prevent reward hacking in Probability Disparity.

\begin{figure*}[htbp]
    \centering
    \includegraphics[width=\textwidth]{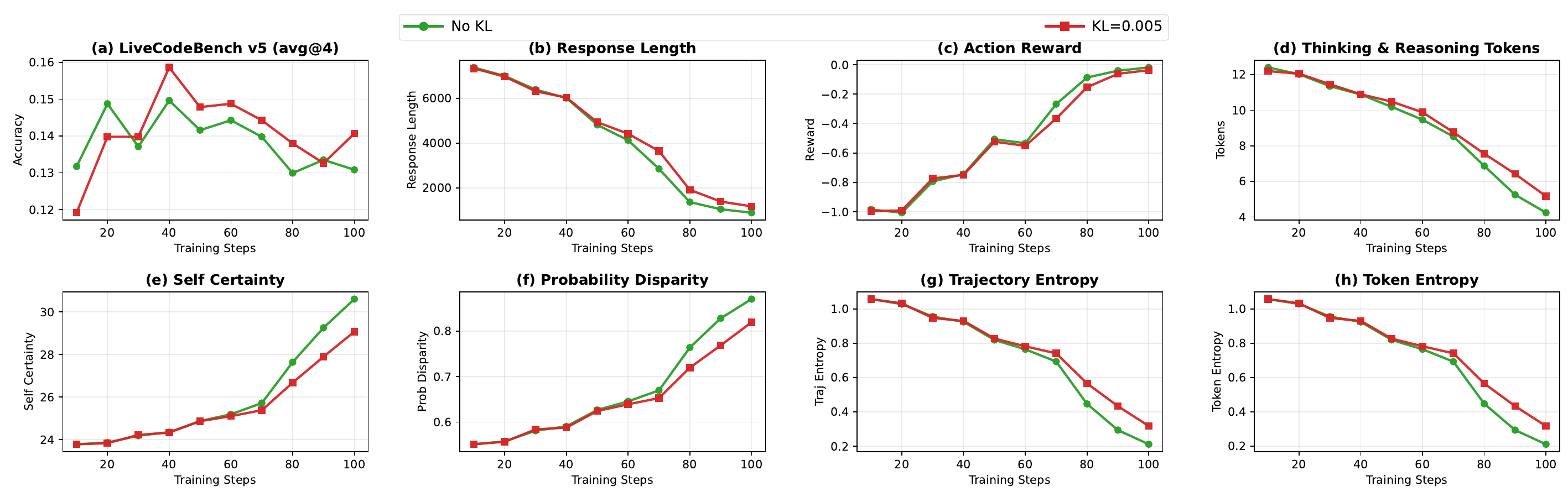}
    \caption{Ablation study on KL coefficient $\beta_{\text{KL}} \in \{0, 0.005\}$ for Probability Disparity. Subplots show (a) LiveCodeBench v5 avg@4 accuracy, (b) response length, (c) action reward, (d) thinking \& reasoning token, (e) self-certainty, (f) probability disparity, (g) trajectory entropy, (h) token entropy.}
    \label{fig:ablation-prob-kl}
\end{figure*}

Figure~\ref{fig:ablation-prob-kl} shows that KL regularization benefits Probability Disparity, analogous to Trajectory-Level Entropy. The $\beta_{\text{KL}}=0.005$ run peaks at 15.9\% (step 40) and achieves a final accuracy of 14.1\%, the best among all Probability Disparity variants, with response lengths at 1{,}170 tokens and trajectory entropy of 0.318, substantially higher than the max-length degenerate baseline (0.005). This indicates that the KL constraint successfully prevents the max-length collapse while preserving meaningful exploration.

\begin{figure*}[htbp]
    \centering
    \includegraphics[width=\textwidth]{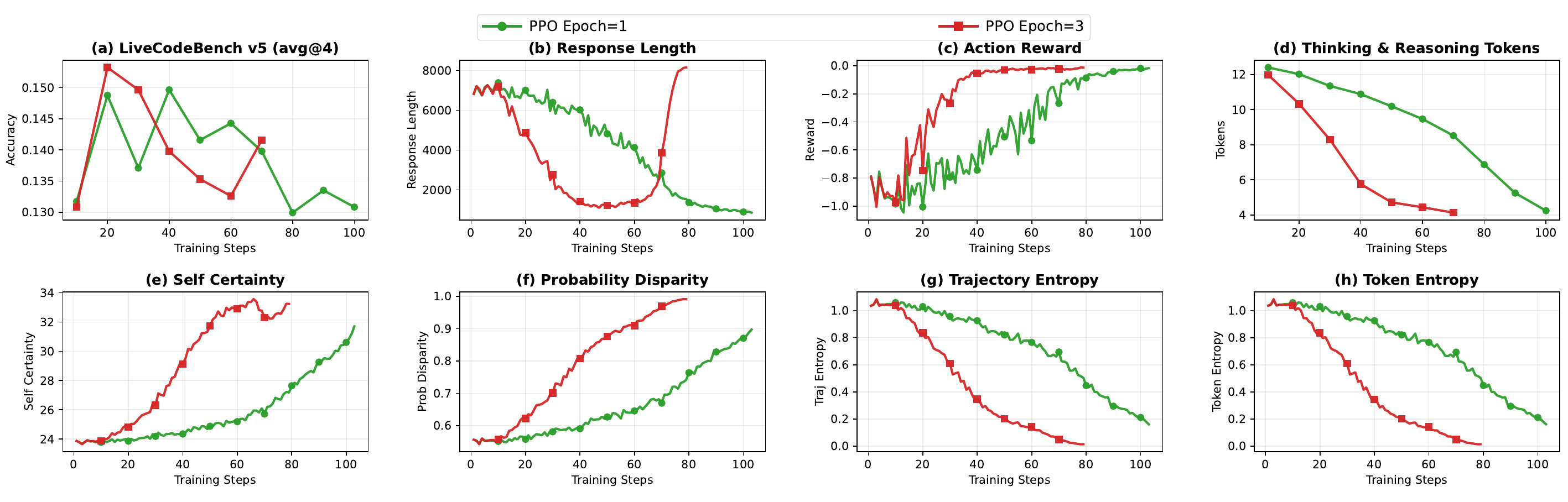}
    \caption{Ablation study on PPO epochs $E \in \{1, 3\}$ for Probability Disparity. Subplots show (a) LiveCodeBench v5 avg@4 accuracy, (b) response length, (c) action reward, (d) thinking \& reasoning token, (e) self-certainty, (f) probability disparity, (g) trajectory entropy, (h) token entropy.}
    \label{fig:ablation-prob-ppo}
\end{figure*}

Figure~\ref{fig:ablation-prob-ppo} shows that $E=3$ induces max-length degenerate behavior for Probability Disparity: response length grows to approximately 8{,}150 tokens with near-zero trajectory entropy (0.014), while accuracy peaks at 15.3\% (step 20) and finishes at 14.2\% at step 79. The max-length collapse indicates the model is exploiting response length rather than improving reasoning, despite the marginally higher accuracy compared to the $E=1$ baseline at the same training window.

\subsubsection{GRPO}
\label{app:ablation_grpo}

\begin{figure*}[htbp]
    \centering
    \includegraphics[width=\textwidth]{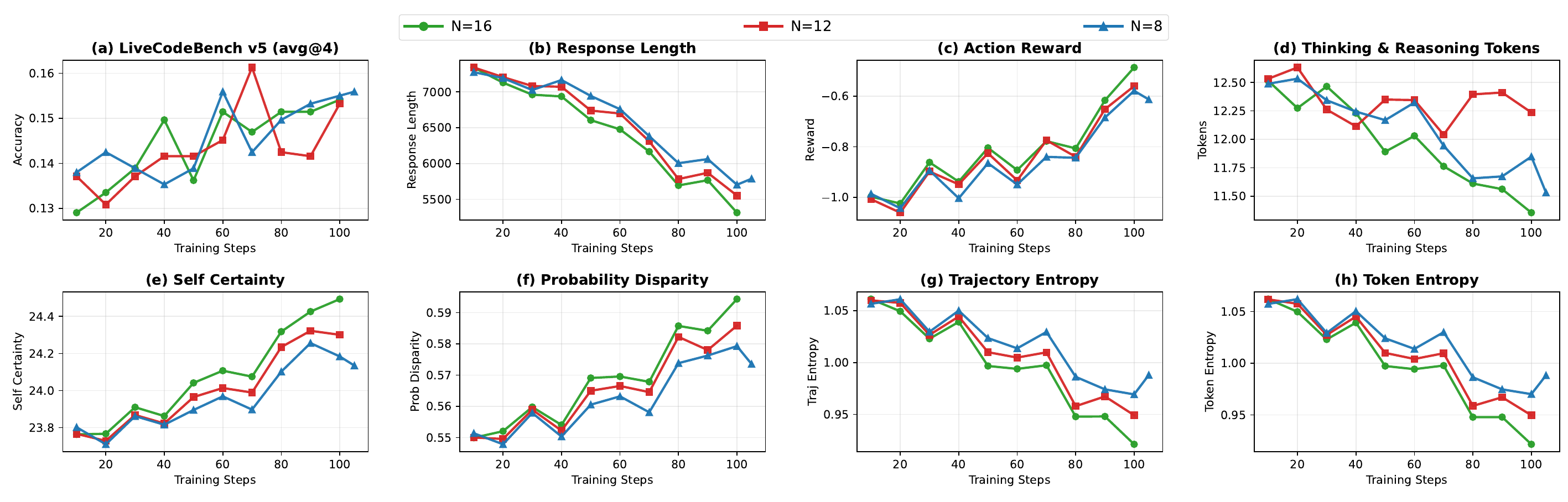}
    \caption{Ablation study on the number of rollouts $N \in \{8, 12, 16\}$ for GRPO. Subplots show (a) LiveCodeBench v5 avg@4 accuracy, (b) response length, (c) action reward, (d) thinking \& reasoning token, (e) self-certainty, (f) probability disparity, (g) trajectory entropy, (h) token entropy.}
    \label{fig:ablation-grpo-num}
\end{figure*}

Figure~\ref{fig:ablation-grpo-num} shows that GRPO is robust to rollout count. All three configurations ($N=8$, $N=12$, $N=16$) maintain high response lengths (5{,}300--5{,}800 tokens) and high trajectory entropy ($\approx$0.92--0.99) throughout training with no collapse observed. $N=12$ achieves the highest peak accuracy at 16.1\% (step 70) and finishes at 15.3\%, while $N=8$ and $N=16$ reach 15.6\% and 15.4\% as final accuracy. Unlike the intrinsic reward methods, GRPO continues improving or remains stable across all rollout counts examined.

\subsubsection{Effect of PPO Epochs Across Methods}
\label{app:ablation_ppo}

\begin{figure*}[htbp]
    \centering
    \includegraphics[width=\textwidth]{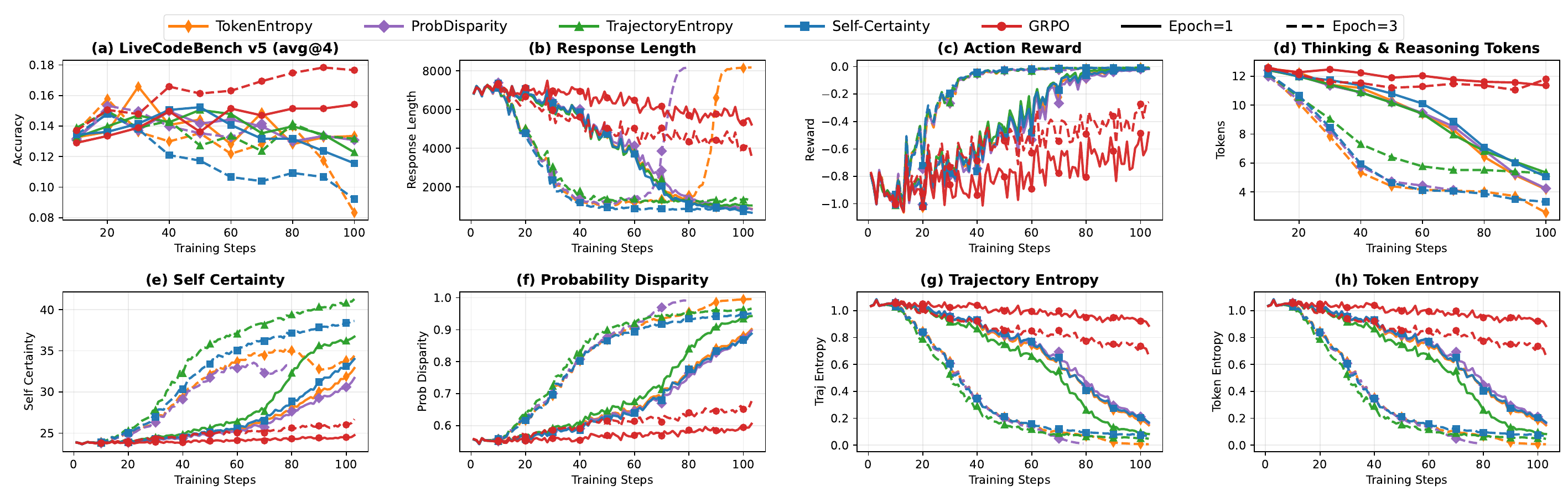}
    \caption{Ablation study on PPO epochs $E \in \{1, 3\}$ across all methods. Subplots show (a) LiveCodeBench v5 avg@4 accuracy, (b) response length, (c) action reward, (d) thinking \& reasoning token, (e) self-certainty, (f) probability disparity, (g) trajectory entropy, (h) token entropy.}
    \label{fig:ablation-all-ppo}
\end{figure*}

Figure~\ref{fig:ablation-all-ppo} presents PPO epoch ablations ($E \in \{1, 3\}$) across all methods simultaneously. The overall pattern is that additional PPO epochs per rollout tend to accelerate or shift the collapse mode of intrinsic reward methods while delivering substantial gains for GRPO (with clip ratio $c=10$).

For Self-Certainty, $E=3$ collapses to short responses at 9.2\% final accuracy, comparable to $E=1$ but reached faster. For Token-Level Entropy, $E=3$ causes drastically faster max-length degenerate collapse (final 8.3\% vs.\ 13.9\% for $E=1$), making it the worst configuration tested. Trajectory-Level Entropy under $E=3$ degrades via short-response collapse to 13.2\%, slightly worse than the $E=1$ baseline. Probability Disparity with $E=3$ drifts into max-length degenerate behavior but retains 14.2\% accuracy within the training window.

In stark contrast, GRPO with $E=3$ and clip ratio $c=10$ achieves the highest accuracy among all PPO-epoch configurations: peaking at 17.8\% (step 90) and maintaining 17.7\% at step 103 with response lengths around 3{,}600 tokens and trajectory entropy of 0.66. This result suggests that for reward-only training without intrinsic bonuses, more PPO epochs combined with appropriate clipping substantially improve sample efficiency without causing collapse.

\section{Scaling Experiments across Model Sizes}
\label{app:scaling}

The main experiments in Section~\ref{sec:experiments} are conducted on DeepSeek-R1-Distill-Qwen-1.5B with an 8k maximum response length.
To examine whether the findings generalize across model scales and architectures, we extend the comparison to two additional models: Qwen3-4B~\citep{yangQwen3TechnicalReport2025} and Qwen2.5-Coder-7B-Instruct~\citep{huiQwen25CoderTechnicalReport2024}.
R1-Distill-1.5B and Qwen3-4B are trained with a 16k maximum response length, while Qwen2.5-Coder-7B uses 8k.
All other hyperparameters follow the default configuration in Section~\ref{sec:experiments}.
We compare five methods---GRPO, Self-Certainty, Token Entropy, Trajectory Entropy, and Probability Disparity---and additionally include a Random reward baseline for R1-Distill-1.5B and Qwen2.5-Coder-7B to assess whether certainty-based rewards provide signal beyond random perturbation.

\subsection{Training Dynamics}
\label{app:scaling_dynamics}

\begin{figure*}[htbp]
    \centering
    \includegraphics[width=\textwidth]{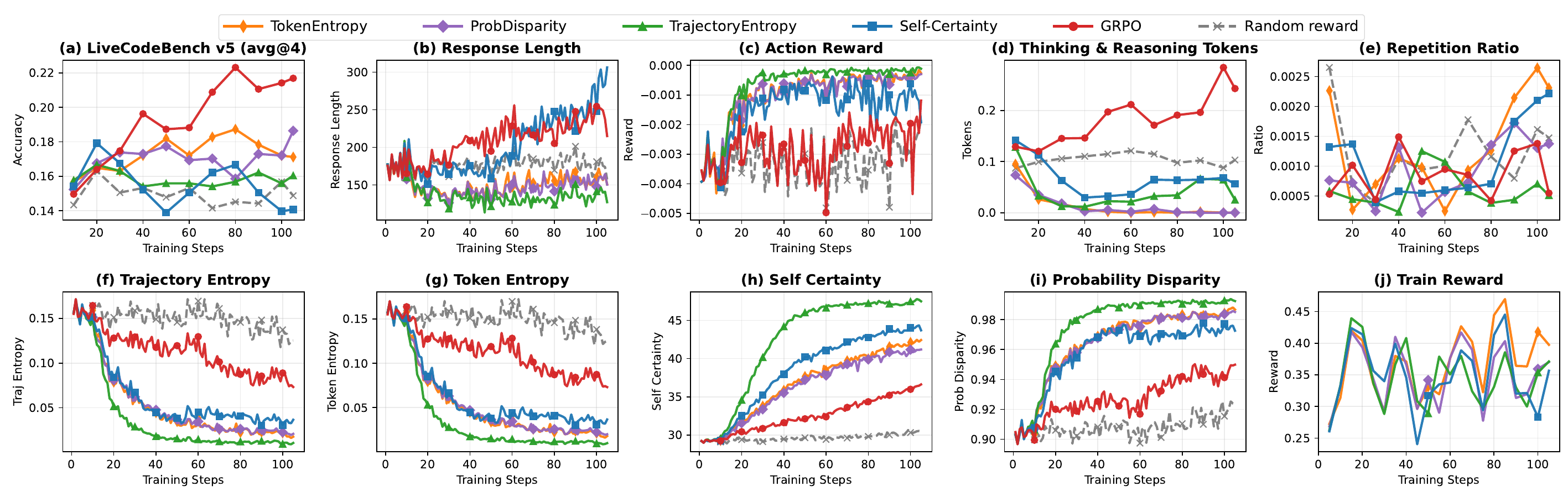}
    \caption{Training dynamics of all methods on Qwen2.5-Coder-7B-Instruct (train 8k) over 105 steps. (a) LiveCodeBench v5 avg@4 accuracy. (b) Response length. (c) Action reward. (d) Thinking and reasoning tokens. (e) Repetition ratio. (f) Training reward. (g) Trajectory entropy. (h) Token entropy. (i) Self-certainty. (j) Probability disparity.}
    \label{fig:scale-main-coder7b}
\end{figure*}

\begin{figure*}[htbp]
    \centering
    \includegraphics[width=\textwidth]{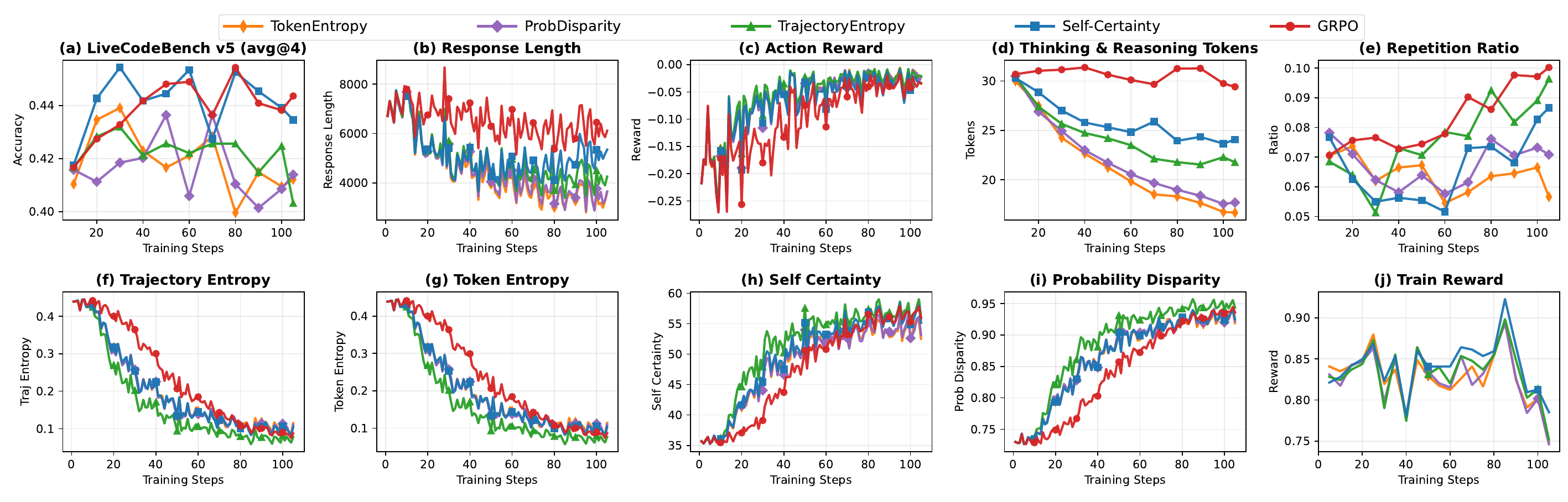}
    \caption{Training dynamics of all methods on Qwen3-4B (train 16k) over 105 steps. (a) LiveCodeBench v5 avg@4 accuracy. (b) Response length. (c) Action reward. (d) Thinking and reasoning tokens. (e) Repetition ratio. (f) Training reward. (g) Trajectory entropy. (h) Token entropy. (i) Self-certainty. (j) Probability disparity.}
    \label{fig:scale-main-qwen3-4b}
\end{figure*}

\begin{figure*}[htbp]
    \centering
    \includegraphics[width=\textwidth]{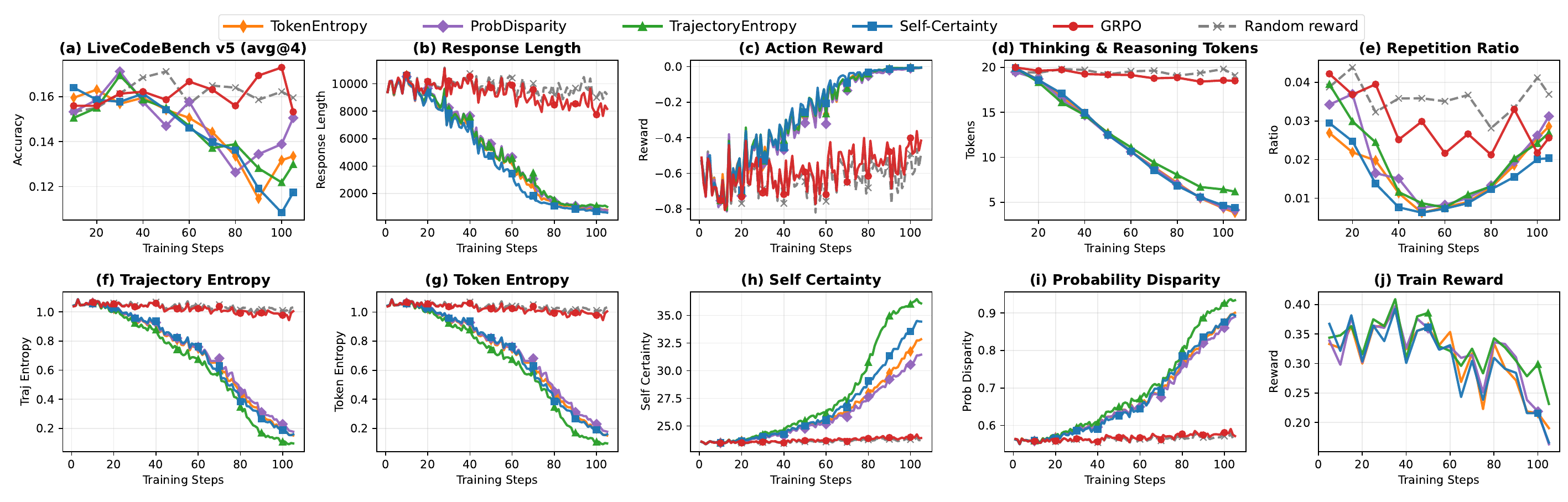}
    \caption{Training dynamics of all methods on R1-Distill-Qwen-1.5B (train 16k) over 105 steps. (a) LiveCodeBench v5 avg@4 accuracy. (b) Response length. (c) Action reward. (d) Thinking and reasoning tokens. (e) Repetition ratio. (f) Training reward. (g) Trajectory entropy. (h) Token entropy. (i) Self-certainty. (j) Probability disparity.}
    \label{fig:scale-main-r1-1.5b}
\end{figure*}

Figures~\ref{fig:scale-main-coder7b}, \ref{fig:scale-main-qwen3-4b}, and \ref{fig:scale-main-r1-1.5b} present the training dynamics for all three model scales, tracking the same ten metrics as Figure~\ref{fig:main_training_dynamics}: LiveCodeBench v5 accuracy, response length, action reward, thinking and reasoning tokens, repetition ratio, training reward, trajectory entropy, token entropy, self-certainty, and probability disparity.

The response length collapse identified in Section~\ref{sec:training_dynamics} persists across all three model scales: certainty-based methods progressively shorten their responses while GRPO maintains stable lengths throughout training.
However, the severity varies with model capacity.
For Qwen2.5-Coder-7B, all four RLIF methods exhibit similar collapse trajectories, with response lengths declining steadily and repetition ratios spiking, closely mirroring the 1.5B patterns.
For Qwen3-4B, the collapse is less uniform: Self-Certainty and Token Entropy maintain comparatively longer responses than Trajectory Entropy and Probability Disparity, consistent with their stronger evaluation performance at this scale (Section~\ref{app:scaling_eval}).
For R1-Distill-1.5B, the patterns are consistent with the findings reported in Section~\ref{sec:training_dynamics}, with all certainty methods converging to short, overconfident outputs.

Across all three scales, trajectory entropy for certainty methods drops toward zero while GRPO maintains values above 0.8, indicating that diversity collapse is a fundamental property of certainty-based optimization rather than a scale-specific artifact.
The thinking and reasoning token count decreases correspondingly for all certainty methods, confirming that the collapse eliminates extended reasoning chains regardless of model capacity.

\subsection{Evaluation Performance}
\label{app:scaling_eval}

\begin{figure*}[htbp]
    \centering
    \includegraphics[width=\textwidth]{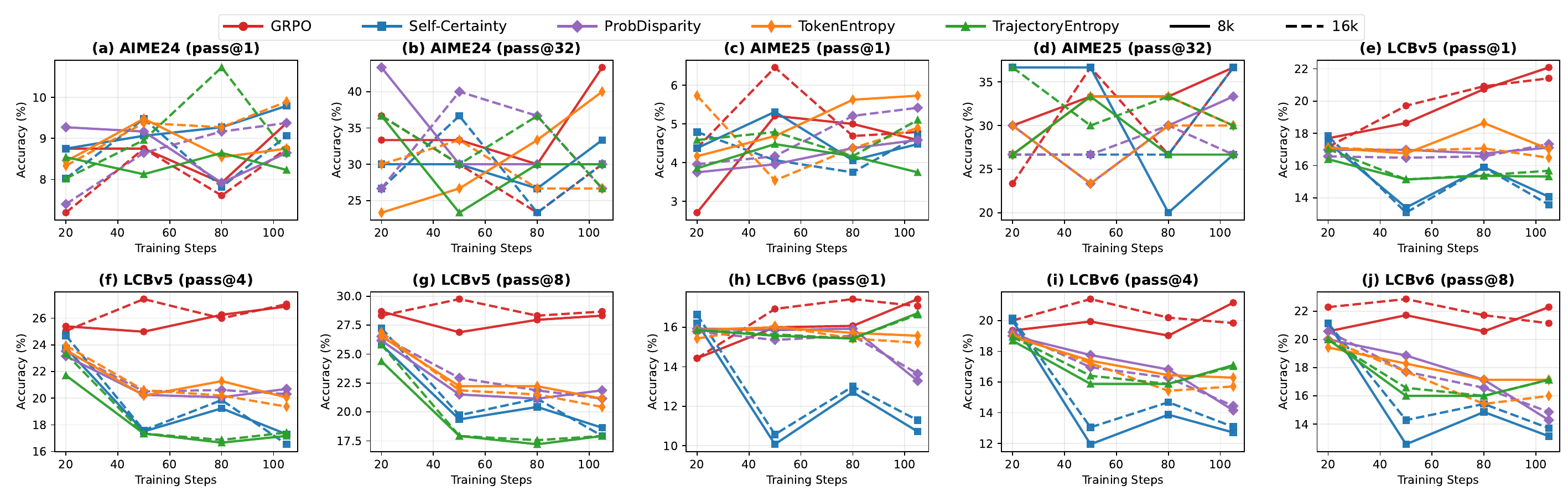}
    \caption{Performance on mathematical reasoning (AIME24, AIME25) and code generation (LiveCodeBench v5, v6) benchmarks throughout training for Qwen2.5-Coder-7B-Instruct (train 8k). Solid and dashed lines denote 8k and 16k maximum generation lengths in evaluation, respectively.}
    \label{fig:scale-eval-coder7b}
\end{figure*}

\begin{figure*}[htbp]
    \centering
    \includegraphics[width=\textwidth]{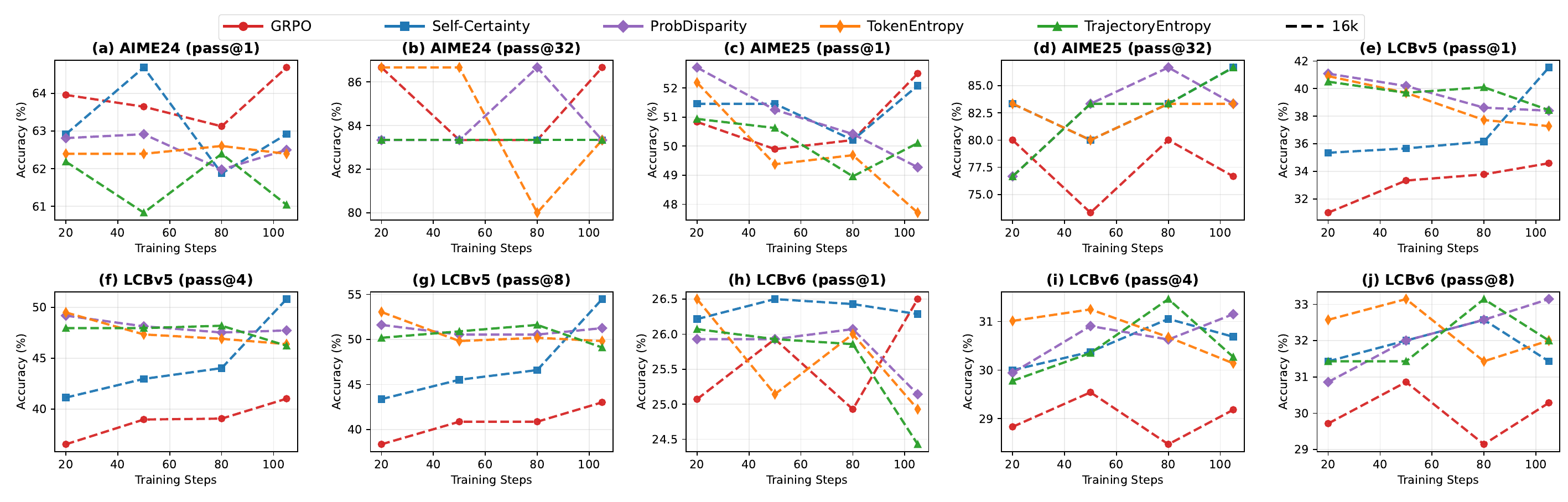}
    \caption{Performance on mathematical reasoning (AIME24, AIME25) and code generation (LiveCodeBench v5, v6) benchmarks throughout training for Qwen3-4B (train 16k). Dashed lines denote 16k maximum generation length in evaluation.}
    \label{fig:scale-eval-qwen3-4b}
\end{figure*}

\begin{figure*}[htbp]
    \centering
    \includegraphics[width=\textwidth]{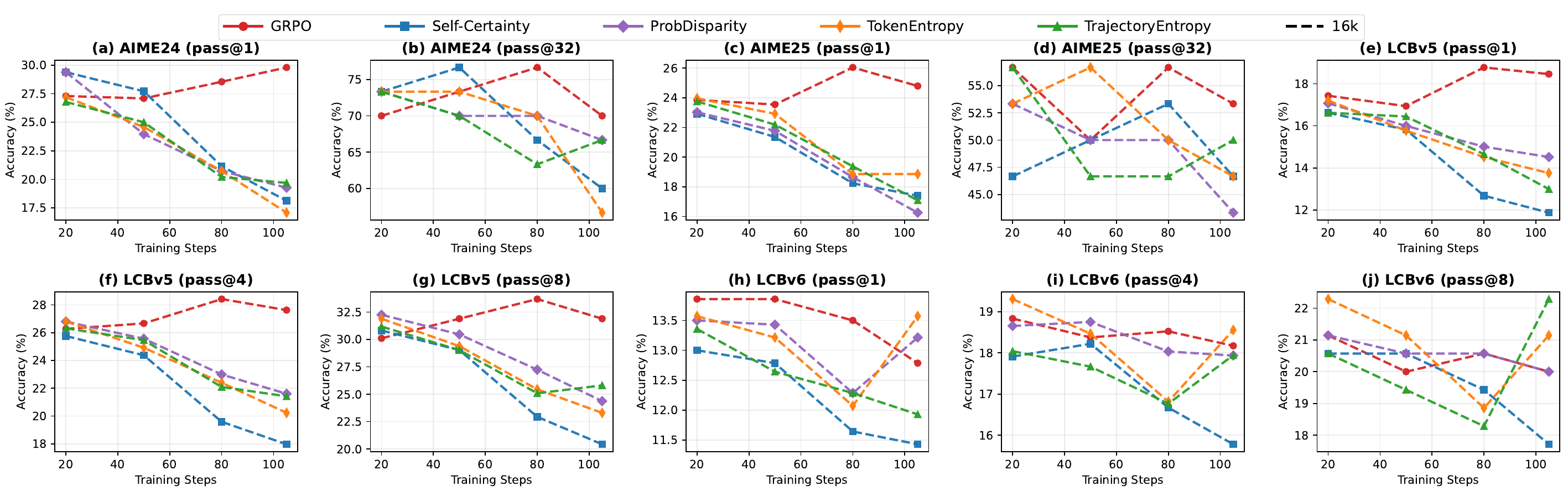}
    \caption{Performance on mathematical reasoning (AIME24, AIME25) and code generation (LiveCodeBench v5, v6) benchmarks throughout training for R1-Distill-Qwen-1.5B (train 16k). Dashed lines denote 16k maximum generation length in evaluation.}
    \label{fig:scale-eval-r1-1.5b}
\end{figure*}

Figures~\ref{fig:scale-eval-coder7b}, \ref{fig:scale-eval-qwen3-4b}, and \ref{fig:scale-eval-r1-1.5b} present pass@$k$ evaluation curves throughout training on code generation (LiveCodeBench v5, v6) and mathematical reasoning (AIME24, AIME25) benchmarks.

\paragraph{R1-Distill-1.5B (train 16k).}
With a 16k training budget, the results are consistent with the 8k findings in Section~\ref{sec:main_results}.
GRPO achieves the highest pass@1 on all four benchmarks (18.5\% LCBv5, 12.8\% LCBv6, 29.8\% AIME24, 24.8\% AIME25), while certainty-based methods trail substantially on code generation, with the best RLIF method reaching only 14.5\% on LCBv5 (Probability Disparity).
On mathematical reasoning, the gap narrows but GRPO retains a clear lead.
Notably, the Random reward baseline achieves 15.2\% on LCBv5 and 27.5\% on AIME24, outperforming all four RLIF methods on AIME.
This suggests that certainty-based optimization actively degrades cross-domain transfer relative to even random perturbation, consistent with the reward hacking dynamics observed in the training curves.

\paragraph{Qwen3-4B (train 16k).}
The 4B scale reveals a qualitatively different pattern.
Self-Certainty achieves the highest pass@1 on LCBv5 at 41.5\%, substantially surpassing GRPO (34.6\%) and Token Entropy (37.3\%).
On LCBv6, Self-Certainty (26.3\%) and GRPO (26.5\%) perform comparably, while on AIME benchmarks GRPO retains a moderate advantage (64.7\% vs.\ 62.9\% AIME24, 52.5\% vs.\ 52.1\% AIME25 for GRPO vs.\ Self-Certainty).
This result contrasts with both the 1.5B and 7B findings where GRPO consistently dominates, indicating that at sufficient model capacity, certainty-based rewards can provide effective in-domain training signal without the severe collapse that undermines smaller models.
The evaluation curves show that Self-Certainty's advantage emerges early in training and persists through step 105, rather than being a transient early-training phenomenon as observed at the 1.5B scale.

\paragraph{Qwen2.5-Coder-7B (train 8k).}
For the code-specialized 7B model trained with an 8k response budget, we report results at both 8k and 16k evaluation lengths.
At 8k evaluation (matching the training length), GRPO leads on code benchmarks (22.1\% LCBv5, 17.4\% LCBv6), with Probability Disparity as the closest RLIF method (17.1\% LCBv5, 13.3\% LCBv6).
Extending evaluation to 16k yields comparable results (GRPO 21.4\%/17.1\%), indicating that the 8k-trained models do not substantially benefit from longer generation budgets.
On mathematical reasoning, all methods perform poorly (below 10\% AIME24 pass@1), reflecting the code-centric pretraining of Qwen2.5-Coder, which limits cross-domain transfer regardless of the training method.
The Random reward baseline (15.5\% LCBv5, 8.4\% AIME24 at 16k eval) performs comparably to RLIF methods, consistent with the 1.5B finding that certainty-based rewards do not reliably outperform random rewards.

\subsection{KL Penalty Ablation across Scales}
\label{app:scaling_kl}

The KL penalty ablation in Appendix~\ref{app:ablation} demonstrates that KL regularization ($\beta_{\text{KL}}=0.005$) moderately mitigates collapse for R1-Distill-1.5B.
A natural question is whether this benefit extends to larger model scales.
We examine KL regularization for all four RLIF methods on Qwen2.5-Coder-7B and R1-Distill-1.5B.

\begin{figure*}[htbp]
    \centering
    \includegraphics[width=\textwidth]{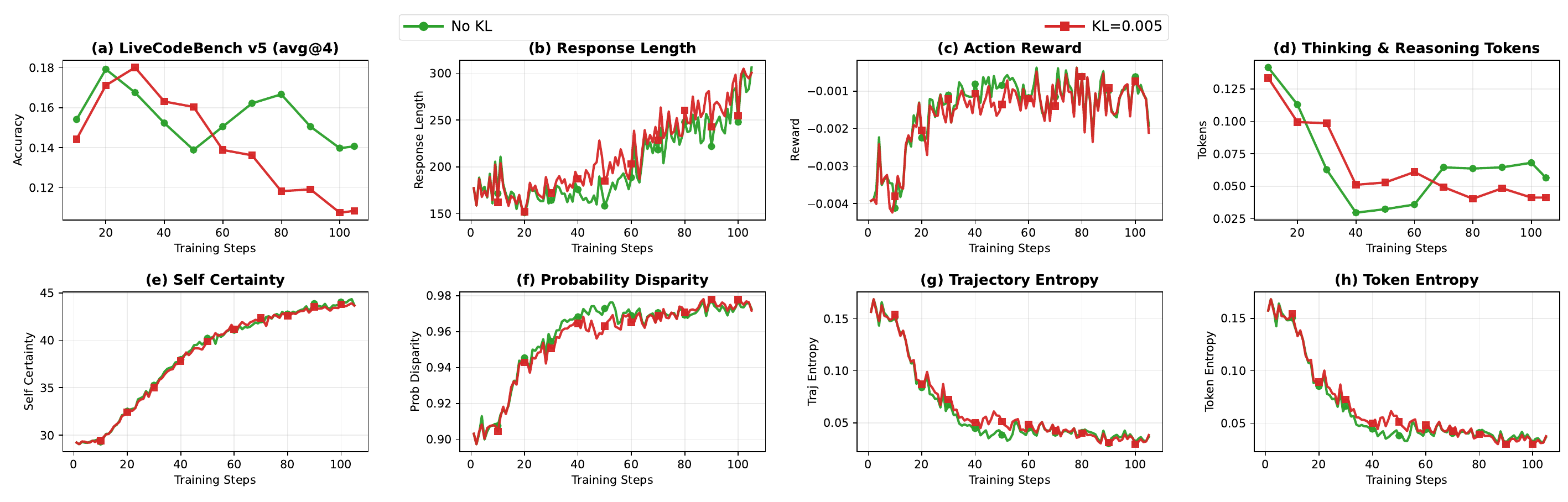}
    \caption{Ablation study on KL coefficient $\beta_{\text{KL}} \in \{0, 0.005\}$ for Self-Certainty on Qwen2.5-Coder-7B-Instruct. (a) LiveCodeBench v5 avg@4 accuracy. (b) Response length. (c) Action reward. (d) Thinking and reasoning tokens. (e) Self-certainty. (f) Probability disparity. (g) Trajectory entropy. (h) Token entropy.}
    \label{fig:scale-kl-coder7b-in}
\end{figure*}

\begin{figure*}[htbp]
    \centering
    \includegraphics[width=\textwidth]{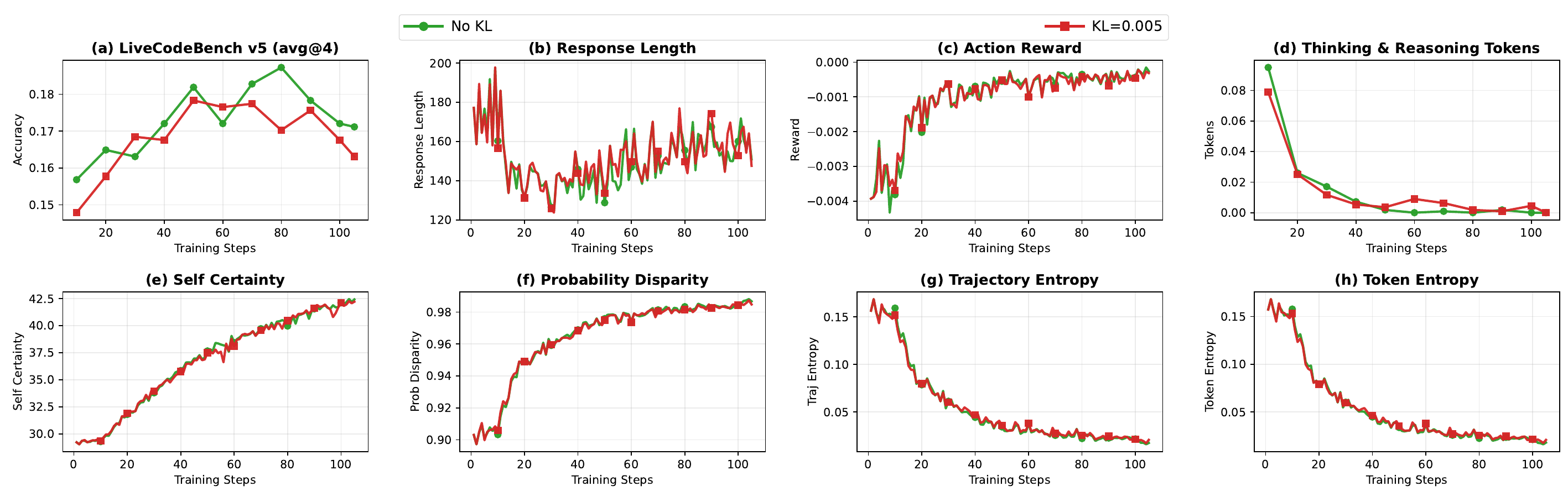}
    \caption{Ablation study on KL coefficient $\beta_{\text{KL}} \in \{0, 0.005\}$ for Token Entropy on Qwen2.5-Coder-7B-Instruct. (a) LiveCodeBench v5 avg@4 accuracy. (b) Response length. (c) Action reward. (d) Thinking and reasoning tokens. (e) Self-certainty. (f) Probability disparity. (g) Trajectory entropy. (h) Token entropy.}
    \label{fig:scale-kl-coder7b-tok}
\end{figure*}

\begin{figure*}[htbp]
    \centering
    \includegraphics[width=\textwidth]{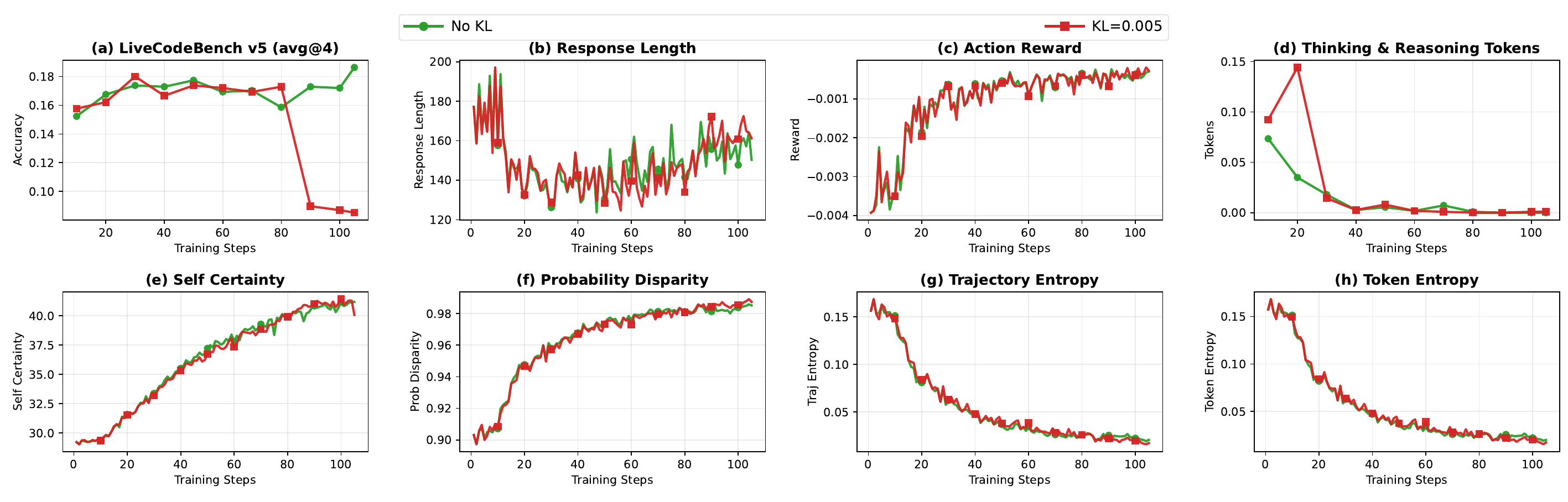}
    \caption{Ablation study on KL coefficient $\beta_{\text{KL}} \in \{0, 0.005\}$ for Probability Disparity on Qwen2.5-Coder-7B-Instruct. (a) LiveCodeBench v5 avg@4 accuracy. (b) Response length. (c) Action reward. (d) Thinking and reasoning tokens. (e) Self-certainty. (f) Probability disparity. (g) Trajectory entropy. (h) Token entropy.}
    \label{fig:scale-kl-coder7b-prob}
\end{figure*}

\begin{figure*}[htbp]
    \centering
    \includegraphics[width=\textwidth]{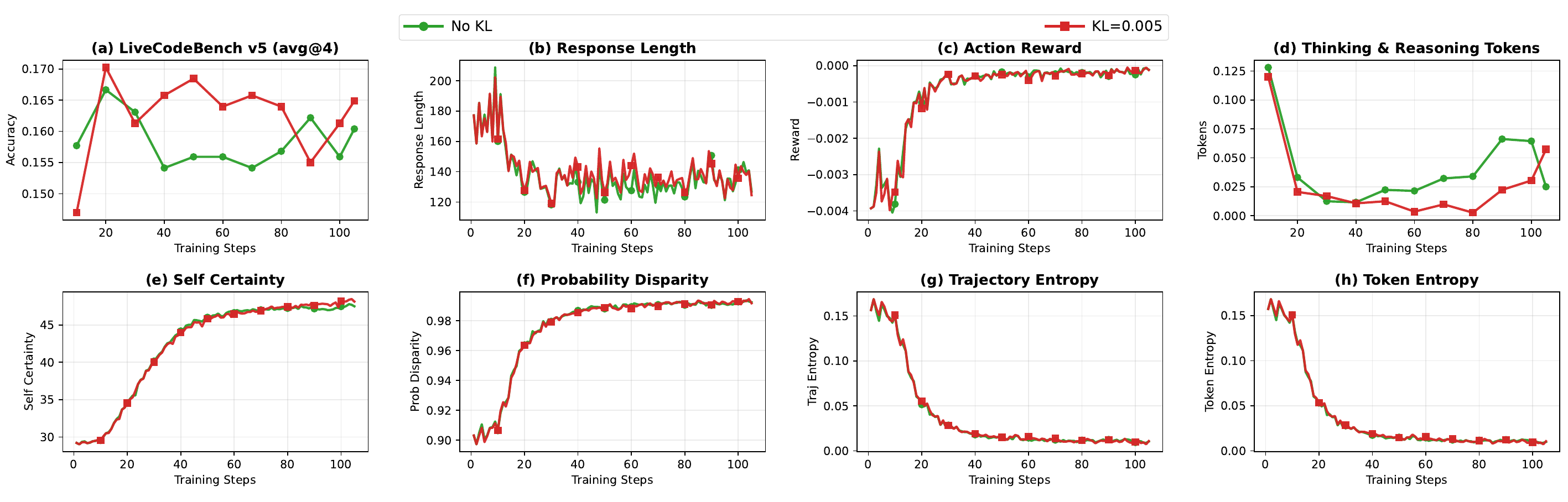}
    \caption{Ablation study on KL coefficient $\beta_{\text{KL}} \in \{0, 0.005\}$ for Trajectory Entropy on Qwen2.5-Coder-7B-Instruct. (a) LiveCodeBench v5 avg@4 accuracy. (b) Response length. (c) Action reward. (d) Thinking and reasoning tokens. (e) Self-certainty. (f) Probability disparity. (g) Trajectory entropy. (h) Token entropy.}
    \label{fig:scale-kl-coder7b-traj}
\end{figure*}

\paragraph{Qwen2.5-Coder-7B.}
Figures~\ref{fig:scale-kl-coder7b-in}--\ref{fig:scale-kl-coder7b-traj} present the training dynamics with and without KL penalty for all four certainty methods on Qwen2.5-Coder-7B.
For Self-Certainty (Figure~\ref{fig:scale-kl-coder7b-in}), the KL-regularized run maintains higher trajectory entropy and longer responses throughout training, slowing the collapse relative to the baseline.
Token Entropy (Figure~\ref{fig:scale-kl-coder7b-tok}) shows a similar pattern, with KL regularization preventing the max-length degenerate collapse observed in the 1.5B $N{=}16$ baseline (Appendix~\ref{app:ablation_token}) and converting it to the more common short-response form.
For Probability Disparity (Figure~\ref{fig:scale-kl-coder7b-prob}) and Trajectory Entropy (Figure~\ref{fig:scale-kl-coder7b-traj}), KL regularization provides consistent but moderate benefits, preserving higher trajectory entropy while not substantially improving final accuracy.
Across all four methods, the KL constraint slows certainty inflation and entropy reduction at the 7B scale, confirming that this effect is not specific to the 1.5B model.

\begin{figure*}[htbp]
    \centering
    \includegraphics[width=\textwidth]{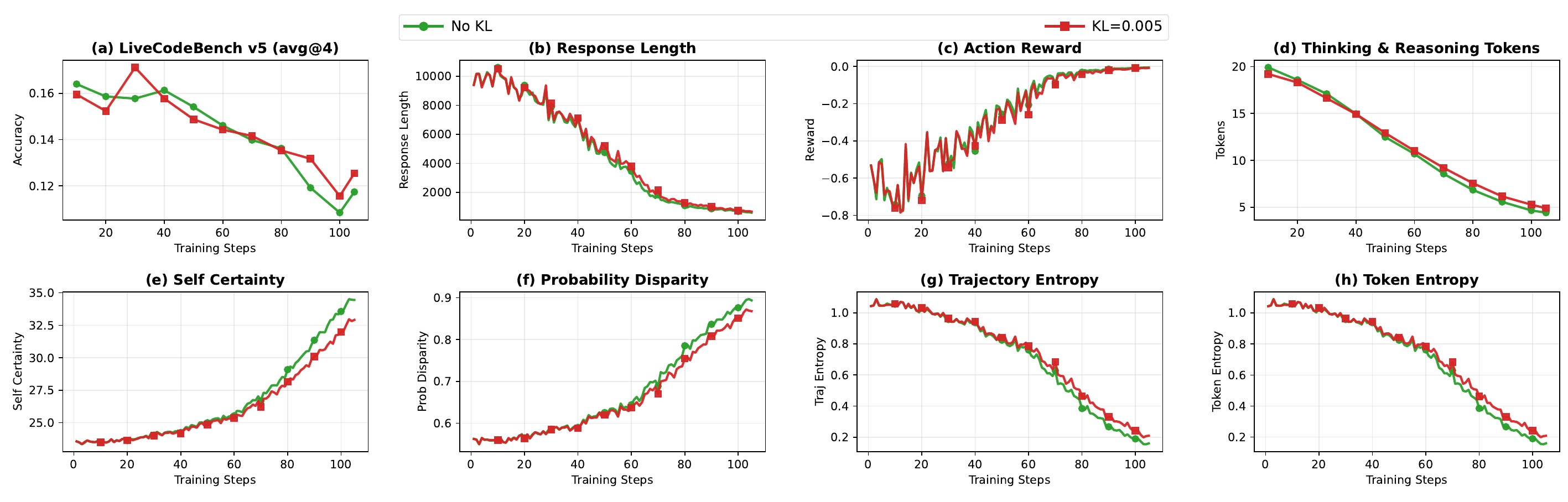}
    \caption{Ablation study on KL coefficient $\beta_{\text{KL}} \in \{0, 0.005\}$ for Self-Certainty on R1-Distill-Qwen-1.5B. (a) LiveCodeBench v5 avg@4 accuracy. (b) Response length. (c) Action reward. (d) Thinking and reasoning tokens. (e) Self-certainty. (f) Probability disparity. (g) Trajectory entropy. (h) Token entropy.}
    \label{fig:scale-kl-r1-1.5b-in}
\end{figure*}

\begin{figure*}[htbp]
    \centering
    \includegraphics[width=\textwidth]{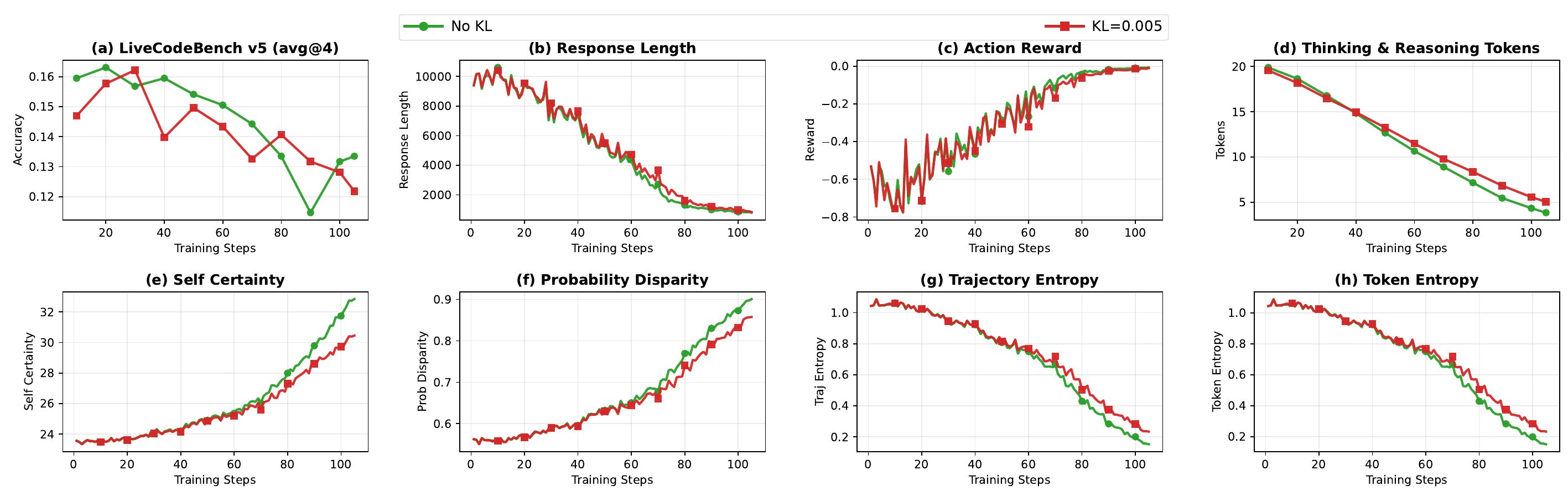}
    \caption{Ablation study on KL coefficient $\beta_{\text{KL}} \in \{0, 0.005\}$ for Token Entropy on R1-Distill-Qwen-1.5B. (a) LiveCodeBench v5 avg@4 accuracy. (b) Response length. (c) Action reward. (d) Thinking and reasoning tokens. (e) Self-certainty. (f) Probability disparity. (g) Trajectory entropy. (h) Token entropy.}
    \label{fig:scale-kl-r1-1.5b-tok}
\end{figure*}

\begin{figure*}[htbp]
    \centering
    \includegraphics[width=\textwidth]{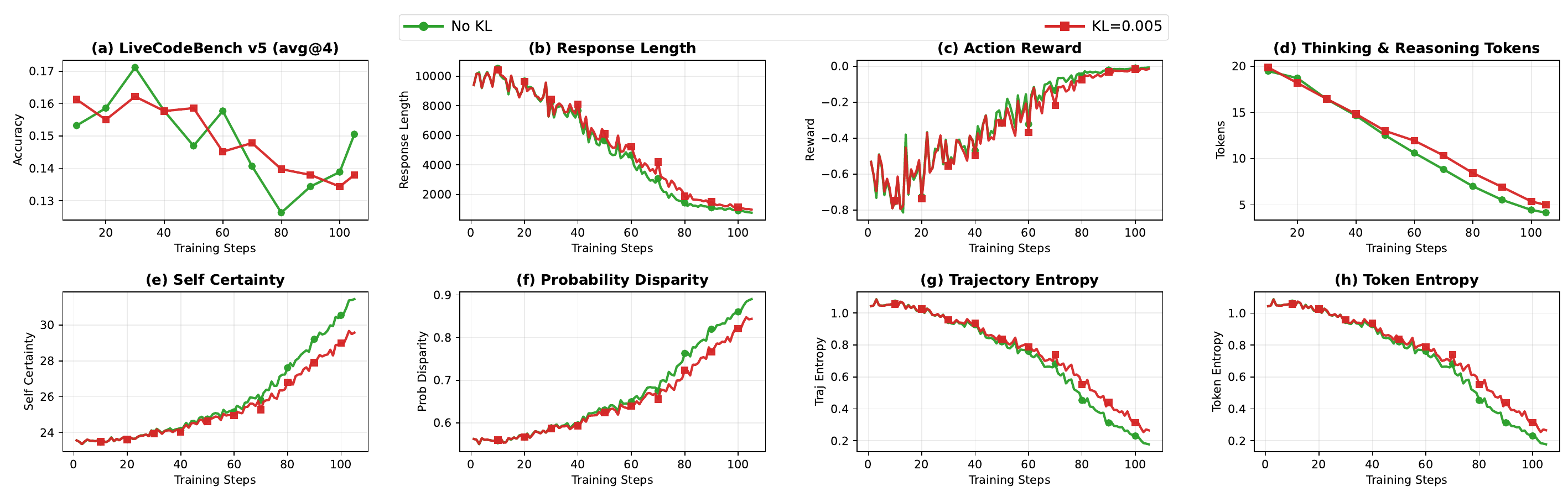}
    \caption{Ablation study on KL coefficient $\beta_{\text{KL}} \in \{0, 0.005\}$ for Probability Disparity on R1-Distill-Qwen-1.5B. (a) LiveCodeBench v5 avg@4 accuracy. (b) Response length. (c) Action reward. (d) Thinking and reasoning tokens. (e) Self-certainty. (f) Probability disparity. (g) Trajectory entropy. (h) Token entropy.}
    \label{fig:scale-kl-r1-1.5b-prob}
\end{figure*}

\begin{figure*}[htbp]
    \centering
    \includegraphics[width=\textwidth]{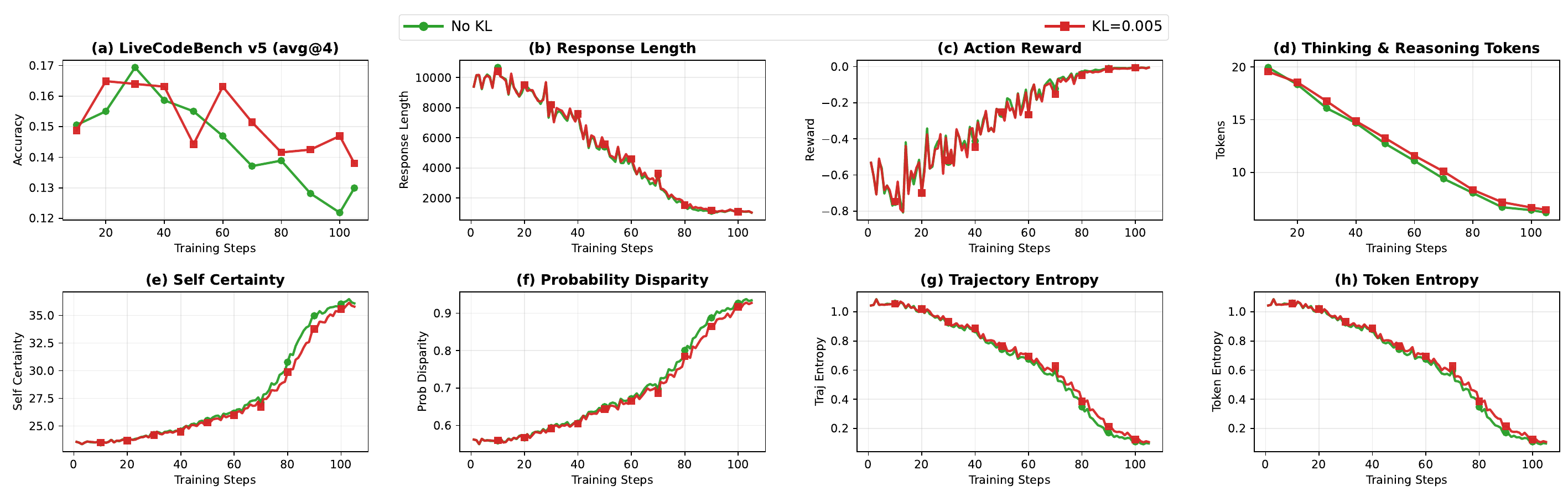}
    \caption{Ablation study on KL coefficient $\beta_{\text{KL}} \in \{0, 0.005\}$ for Trajectory Entropy on R1-Distill-Qwen-1.5B. (a) LiveCodeBench v5 avg@4 accuracy. (b) Response length. (c) Action reward. (d) Thinking and reasoning tokens. (e) Self-certainty. (f) Probability disparity. (g) Trajectory entropy. (h) Token entropy.}
    \label{fig:scale-kl-r1-1.5b-traj}
\end{figure*}

\paragraph{R1-Distill-1.5B.}
Figures~\ref{fig:scale-kl-r1-1.5b-in}--\ref{fig:scale-kl-r1-1.5b-traj} present the corresponding results for R1-Distill-1.5B.
The patterns are consistent with the 8k KL ablation in Figure~\ref{fig:ablation-intuitor-kl}: KL regularization moderately slows response length collapse for Self-Certainty (Figure~\ref{fig:scale-kl-r1-1.5b-in}), with the regularized run retaining longer responses and higher trajectory entropy at the end of training.
Token Entropy (Figure~\ref{fig:scale-kl-r1-1.5b-tok}), Probability Disparity (Figure~\ref{fig:scale-kl-r1-1.5b-prob}), and Trajectory Entropy (Figure~\ref{fig:scale-kl-r1-1.5b-traj}) show similar trends, with KL regularization delaying but not preventing the underlying collapse.

\end{document}